\documentclass{article}

\usepackage[preprint]{neurips_2026}
\usepackage[utf8]{inputenc}
\usepackage[T1]{fontenc}
\usepackage{amsmath,amssymb}
\usepackage{graphicx}
\usepackage{booktabs}
\usepackage{multirow}
\usepackage{xcolor}
\usepackage{hyperref}
\usepackage{cleveref}
\usepackage{subcaption}
\usepackage{microtype}
\usepackage[font=small,labelfont=bf,skip=4pt]{caption}
\usepackage{tikz}
\usepackage{pgfplots}
\pgfplotsset{compat=1.18}
\usepgfplotslibrary{groupplots}
\usetikzlibrary{calc}

\title{Decomposing the Depth Profile of Fine-Tuning\thanks{All experiment code, hyperparameters, and reference results for the 240 runs are available at \url{https://github.com/jb1999/ft-depth-profile-paper}. The repository includes scripts to reproduce all primary and scale-verification runs, per-model depth profile tables, and instructions for replicating the equal-step control.}}

\author{Jayadev Billa\thanks{Unaffiliated researcher; previously at ISI@USC, Yahoo, Nuance, and BBN.} \\
\texttt{jbilla2004@gmail.com}}

\begin{document}
\maketitle

\begin{abstract}
Fine-tuning adapts pretrained networks to new objectives. Whether the resulting depth profile of representational change reflects an intrinsic property of the model or the magnitude of gradient flow has not been tested directly. We measure this profile across 240 fine-tuning runs spanning 15 models in four architecture families (encoder and decoder transformers, a state-space model, and an RNN) at scales from 125M to 6.9B parameters. Representational change concentrates in output-proximal layers in every standard-training run except one. We apply a per-layer control that equalizes $\|\Delta W\|/\|W\|$ across layers after each optimizer step. Under this control, the profile persists in some conditions and collapses in others. At 125M--350M, sequential-block architectures (BERT, OPT, GPT-2) retain the slope across tested objectives while parallel-block architectures (Pythia, CodeGen) retain it only for causal-language-modeling objectives. This architectural distinction narrows at 1.3B--1.4B, where both block types show positive equal-step slopes for CausalLM. Under standard training, profile shape is described by two additional axes: steepness tracks a training-free objective distance at initialization, and profile width is dominated by architecture. We treat the locality gradient, the depthwise slope of representational change, as a composite phenomenon whose components are scale-dependent.
\end{abstract}

\section{Introduction}
\label{sec:intro}

Current fine-tuning practice relies on depth-allocation heuristics. Common examples include freezing lower layers, placing LoRA modules \citep{hu2022lora} or adapters \citep{houlsby2019parameter} late in the network, and assigning discriminative learning rates by depth \citep{howard2018universal}. Each of these choices assumes that some layers change more than others during fine-tuning, and that the pattern is consistent enough to exploit. Whether it holds across architectures and training objectives, what controls its shape, and whether it reflects representational structure or merely the magnitude of gradient flow, have not been studied systematically.

Prior work on representational change during fine-tuning has three limitations. It is overwhelmingly transformer-specific: the observation that later layers change more has been tested almost exclusively on BERT-family models \citep{merchant2020what,zhou2022closer,durrani2021transfer}. It is largely qualitative, reporting the direction of concentration but not quantifying its magnitude or how it varies. And it does not address the most basic question about the pattern's origin: later layers are closer to the loss and receive larger gradient updates through backpropagation. A monotonically decreasing profile of change with distance from the output could simply reflect this gradient-magnitude asymmetry rather than any intrinsic property of the representation.

In this paper, we measure the depthwise distribution of representational change across a broader set of conditions than has been reported before, and we use a targeted optimizer control to decompose the observed profile into components that do and do not depend on gradient-magnitude asymmetry. We evaluate nine models across four architecture families (encoder and decoder transformers, a state-space model, and an RNN) for the primary analysis, with scale verification on six additional decoder models from 1.3B to 6.9B parameters, totaling 240 fine-tuning runs across 15 models. For each run we compute the change in representations at seven relative depths using Procrustes distance and linear CKA. We introduce a training-free measure of the distance between two training objectives, computed from the representation gradient at initialization. And we apply a per-layer trust-ratio control (equal-step) that normalizes $\|\Delta W\|/\|W\|$ across all layer groups after each optimizer step; this is the same per-layer ratio used in LARS \citep{you2017lars} and LAMB \citep{you2020lamb}, applied here as a diagnostic rather than as an optimizer mechanism.

Our results show that the locality gradient (the slope of representational change with layer depth, defined formally in \Cref{sec:setup}) is robust under standard training: in 189 of 190 standard-training runs, representational change increases toward the output. The equal-step control then shows that this pattern is not a single phenomenon. In architectures with sequential residual blocks, where attention output feeds into the FFN (BERT, OPT, GPT-2), the slope survives or amplifies when per-layer update magnitudes are equalized across the tested objectives. In architectures with parallel residual blocks, where attention and MLP read the same residual state and their outputs are summed (Pythia, CodeGen), survival is selective: causal-language-modeling-family objectives retain a positive slope, contrastive objectives do not. The contrastive-collapse pattern is consistent across the two parallel-block decoders and also affects SimCSE in some sequential decoders, suggesting that contrastive-objective fragility under equal-step reflects an objective-driven property, while the architecture distinction acts mainly on causal-LM-family objectives. We therefore treat the observed locality gradient as a composite, not a single mechanism.

Beyond the decomposition, two axes describe the shape of the profile under standard training. Its steepness tracks the training-free objective distance: under gradient Procrustes, 8 of 9 model-level Spearman correlations between distance and slope are positive ($p \approx 0.039$, sign test); under cosine distance, 9 of 9 are positive ($p \approx 0.004$). Its width is dominated by architecture: the final layer absorbs 42\% of total representational change in BERT-base and 99\% in GPT-2, and cross-architecture variance is several times larger than within-architecture variance. We do not claim a prescriptive PEFT recipe or a closed-form mechanism; we provide an empirical decomposition and regularities that constrain both.

\paragraph{Contributions.}
\begin{itemize}
    \item \textbf{Cross-architecture universality.} The locality gradient holds in 189 of 190 standard-training runs, spanning encoder and decoder transformers, a state-space model, and an RNN. It is not a transformer artifact.
    \item \textbf{An equal-step control revealing selective persistence.} A per-layer trust-ratio condition that equalizes $\|\Delta W\|/\|W\|$ across layers shows the profile is not a single phenomenon: at 125M--350M, sequential-block architectures (BERT, OPT, GPT-2) retain positive slopes across tested objectives (10 of 12), while parallel-block architectures (Pythia, CodeGen) retain them only for causal-LM-family objectives, and contrastive objectives collapse in both parallel-block decoders and some sequential causal decoders. Scale experiments at 1.3B--1.4B show this architectural distinction narrows: both block types show positive CausalLM equal-step slopes at larger scale (\Cref{sec:equalstep}). To our knowledge this is the first direct test of whether output-localized fine-tuning reflects representational structure, gradient magnitude, or their interaction with objective family.
    \item \textbf{A two-axis description of profile shape.} Under standard training, steepness tracks a training-free objective distance computed from the representation gradient at initialization (8 of 9 model-level Spearman correlations positive under gradient Procrustes, $p \approx 0.039$; 9 of 9 under cosine distance, $p \approx 0.004$); width is dominated by architecture.
\end{itemize}

\section{Related Work}
\label{sec:related}

\paragraph{Representational change during fine-tuning.}
\citet{yosinski2014transferable} studied layer transferability in CNNs, showing that early layers learn more general features while later layers are more task-specific. \citet{merchant2020what} tracked changes in BERT representations during fine-tuning and observed that later layers change more; \citet{zhou2022closer} extended this analysis, examining what changes in addition to where. \citet{durrani2021transfer} probed three pretrained transformers and showed that linguistic knowledge, distributed across layers before fine-tuning, becomes localized to lower layers afterward. \citet{neyshabur2020what} identified feature reuse and low-level data statistics as important factors in transfer learning, and \citet{kumar2022finetuning} showed that full fine-tuning can distort pretrained features relative to linear probing on the frozen backbone. \citet{tenney2019bert} showed that BERT's layers recover the classical NLP pipeline in approximate order, and \citet{ramasesh2021anatomy} used CKA to show that deeper layers are disproportionately responsible for catastrophic forgetting. \citet{raghu2021vision} compared vision transformers and CNNs and found architecture-dependent specialization patterns. Our work broadens this line of inquiry: we measure the depthwise change profile across fifteen models (125M--6.9B) spanning transformers, SSMs, and RNNs, and use an equal-step optimizer control to decompose the profile into components that do and do not depend on update-magnitude asymmetry.

\paragraph{Selective adaptation and parameter-efficient fine-tuning.}
\citet{howard2018universal} introduced discriminative learning rates and gradual unfreezing. LoRA \citep{hu2022lora} and adapters \citep{houlsby2019parameter} insert trainable modules at selected layers; \citet{biderman2024lora} measured the layer-by-layer rank of LoRA updates and found that full fine-tuning produces perturbations with $10$--$100$ higher effective rank. BitFit \citep{benzaken2022bitfit} trains only bias terms. \citet{lee2023surgical} showed that the optimal fine-tuning layer depends on the type of distribution shift, and \citet{pan2024lisa} observed that LoRA weight norms are skewed across layers. In each case the decision of \emph{where} to adapt is made heuristically or by post-hoc analysis. \citet{aghajanyan2021intrinsic} showed that fine-tuning operates in a low-dimensional subspace; our work adds the spatial dimension and a test of whether the spatial structure is intrinsic.

\paragraph{Similarity metrics and concept erasure.}
We rely on CKA \citep{kornblith2019similarity} and Procrustes distance \citep{williams2021generalized, schonemann1966generalized} to measure representational change; \citet{davari2023reliability} showed that CKA values can be manipulated without substantial functional changes, motivating our use of both and our verification that they agree. For concept erasure we use LEACE \citep{belrose2023leace}, which provides closed-form linear concept erasure. CCA-based metrics \citep{raghu2017svcca, morcos2018insights} address similar measurement problems.

\section{Setup}
\label{sec:setup}

\subsection{Definitions}

\paragraph{Plasticity profile and locality gradient.} We study a pretrained model with $L$ layers, fine-tuned on a new objective starting from pretrained weights. For each run we sample activations before and after fine-tuning at seven relative depths $d = \ell/L \in \{0.10, 0.25, 0.40, 0.60, 0.75, 0.90, 1.00\}$, using a fixed set of 290 probe inputs. Activations are mean-pooled over non-padding tokens to produce matrices $\mathbf{X}, \mathbf{Y} \in \mathbb{R}^{n \times D}$ ($n$ probe inputs, $D$ hidden dimension). We measure change at each depth using two complementary metrics. \emph{Procrustes distance}: after centering and Frobenius-normalizing both matrices, we compute $\Delta_P = \|\mathbf{X} - \mathbf{Y}\mathbf{R}^*\|_F$, where $\mathbf{R}^*$ is the orthogonal matrix best aligning $\mathbf{Y}$ to $\mathbf{X}$ via SVD \citep{schonemann1966generalized}, used as a distance between representations following \citet{williams2021generalized}. $\Delta_P$ is bounded in $[0, \sqrt{2}]$: zero means identical up to rotation, $\sqrt{2}$ means essentially uncorrelated. \emph{CKA-based change}: $\Delta_{\text{CKA}} = 1 - \text{CKA}(\mathbf{X}, \mathbf{Y})$ using linear CKA \citep{kornblith2019similarity}, bounded in $[0,1]$. We fit a line $\Delta(d) = \alpha d + \beta$ to the seven-point profile; the slope $\alpha$ is the \emph{locality gradient}, positive when change increases toward the output.

\paragraph{Objective distance.} We define a model-conditioned, training-free distance based on the gradient of the loss with respect to the last-layer representation: this is the signal that fine-tuning would propagate backward, and two objectives that drive that signal in similar directions will reshape the representation in similar ways. Alternatives include Fisher-information embeddings \citep{achille2019task2vec} and transfer-performance matrices \citep{standley2020which}; both require either labeled data or multi-task transfer experiments and are substantially more expensive at scale. The reference objective for each model is the pretraining objective (MLM for encoders, CausalLM for causal models), which has distance zero by definition; all other objectives are measured relative to it. Concretely, for each objective $A$ we run a fixed sequence of input batches through the pretrained model and capture $\partial \mathcal{L}_A / \partial \mathbf{h}^{(L)}$ at the final layer via a backward hook, mean-pool over non-padding tokens, and stack across batches to form $\mathbf{G}_A \in \mathbb{R}^{n \times D}$. We repeat for objective $B$ on the same model and same batches to obtain $\mathbf{G}_B$, so rows of $\mathbf{G}_A$ and $\mathbf{G}_B$ correspond to the same input under the same batch context. Our primary distance is the Procrustes distance on the full stacked matrices, $\|\mathbf{G}_A - \mathbf{G}_B \mathbf{R}^*\|_F$; a secondary metric is the cosine dissimilarity of the mean gradient vectors. Details on the batch-context caveat and the split-half reliability of the orderings are in \Cref{app:metrics}.

\subsection{Models, objectives, and training protocol}

\paragraph{Models.} We study fifteen models across four architecture families. The primary analysis uses nine models: \emph{transformer encoders} BERT-base (12 layers, 110M parameters) \citep{devlin2019bert}, BERT-large (24 layers, 340M) \citep{devlin2019bert}, and RoBERTa-base (12 layers, 125M) \citep{liu2019roberta}; \emph{transformer decoders} Pythia-160m (12 layers, 160M) \citep{biderman2023pythia}, OPT-125m (12 layers, 125M) \citep{zhang2022opt}, GPT-2 medium (24 layers, 345M) \citep{radford2019language}, and CodeGen-350M-mono (20 layers, 350M) \citep{nijkamp2023codegen}; a \emph{state-space model} Mamba-130m (24 layers, 130M) \citep{gu2024mamba}; and a \emph{recurrent network} RWKV-4-169m (12 layers, 169M) \citep{peng2023rwkv}. The four decoder transformers differ in residual-block structure: OPT and GPT-2 use sequential blocks (attention output feeds into FFN), while Pythia and CodeGen use parallel blocks (attention and MLP operate on the same input and their outputs are summed). This distinction is central to the equal-step analysis in \Cref{sec:equalstep}. Six larger decoder transformers are included as scale verification: OPT-1.3B, OPT-2.7B, and OPT-6.7B \citep{zhang2022opt}, and Pythia-1.4B, Pythia-2.8B, and Pythia-6.9B \citep{biderman2023pythia}, covering 1.3B to 6.9B parameters. These models follow the same sequential and parallel block designs as their smaller counterparts.

\paragraph{Objectives.} For encoder models we use five objectives: masked language modeling (MLM), next-sentence prediction (NSP, BERT only), span denoising, SimCSE \citep{gao2021simcse}, and Barlow Twins \citep{zbontar2021barlow}. For causal models the corresponding set is causal language modeling, causal span denoising, SimCSE (last-token pooling), and Barlow Twins (last-token pooling). All objectives use the model's pretrained head or no head at all, so no randomly initialized parameters appear in the gradient path.

\paragraph{Training protocol.} Each model-objective pair is trained under up to six optimizer conditions: (1)~standard AdamW with layer-wise learning rate decay, (2)~uniform learning rate, (3)~LoRA adapters at all layers, (4)~frozen LayerNorm parameters, (5)~frozen bottom 50\% of layers, and (6)~equal-step, described below. The equal-step control is applied to the five models used for the block-type analysis (BERT-base, GPT-2, OPT, Pythia, CodeGen). Encoder models are trained for 3 epochs under conditions 1--5 and 1 epoch under equal-step; causal models train for 1 epoch throughout. Batch size is 32 (reduced to 16 for GPT-2 and CodeGen on contrastive objectives) and learning rate is $2\times10^{-5}$. Three-seed validation is conducted for BERT-base and Pythia under the standard condition. The primary analysis spans 9 models, 4--5 objectives per model, and up to 6 optimizer conditions, totaling 226 fine-tuning runs. Scale verification adds 14 runs across 6 additional models (240 total; see \Cref{app:scale_verification}). Equal-step is not applied to RoBERTa, BERT-large, Mamba, or RWKV because the trust-ratio control is defined over transformer layer groups and does not straightforwardly extend to SSM and RNN parameter groupings. The full run accounting is in \Cref{app:training}.

\paragraph{Equal-step condition.} The locality gradient could reflect a trivial consequence of gradient flow: later layers are closer to the loss and receive larger updates. To test this, we introduce a per-layer trust-ratio condition that equalizes update magnitudes across depth. After each optimizer step, we rescale per-layer updates so that $\|\Delta W_\ell\| / \|W_\ell\| = \tau$ for all layer groups $\ell$, where $\tau = 10^{-3}$ is the trust ratio, the ratio of update norm to weight norm per layer. This is the same quantity that LARS \citep{you2017lars} and LAMB \citep{you2020lamb} normalize per layer to stabilize large-batch training: LAMB sets $\|\Delta W_\ell\| / \|W_\ell\|$ equal to a fixed global learning rate for every layer, which is structurally identical to our rescaling. The difference is purpose: LAMB applies the normalization as part of the optimizer to equalize effective step sizes; we apply it post-hoc to a standard AdamW run as a diagnostic, using a fixed $\tau$ chosen to match the representational movement of the standard-training runs (\Cref{app:equalstep}). The rescaling preserves the \emph{direction} of each layer's update while removing the gradient-magnitude asymmetry. Trainable parameters are grouped by transformer layer index (13 groups for 12-layer models: one per layer plus embeddings); the rescaling is applied after the AdamW step and before the next forward pass. If the locality gradient survives this equalization, the depth profile reflects something beyond gradient magnitude. Details are in \Cref{app:equalstep}.

\section{Results}
\label{sec:results}

\subsection{Locality is universal}
\label{sec:universal}

\begin{figure}[t]
\centering
\input{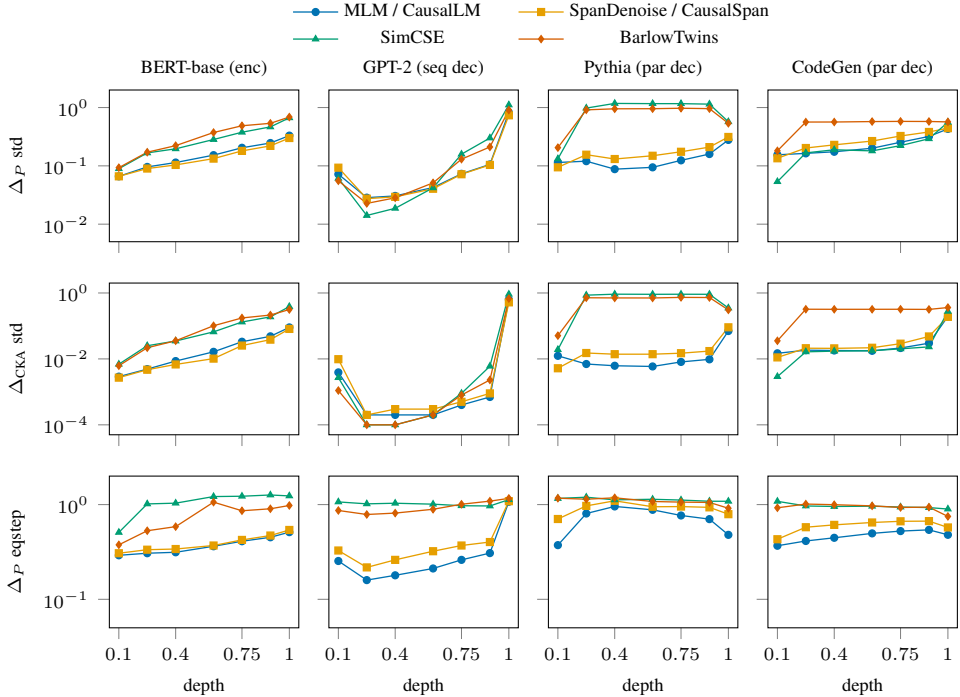}
\caption{Depthwise plasticity profiles for one encoder (BERT-base), one sequential-block decoder (GPT-2), and two parallel-block decoders (Pythia, CodeGen). Rows: Procrustes under standard training, CKA under standard training, Procrustes under equal-step. $y$-axes log-scaled. Non-transformer profiles (Mamba, RWKV) and full per-model tables are in \Cref{app:profiles}.}
\label{fig:depth_profiles}
\end{figure}

\Cref{fig:depth_profiles} shows representative depthwise profiles for several model-objective pairs. In every case, representational change is small at early layers and increases toward the output. The shape is consistent across all four architecture families, though the steepness and concentration differ (discussed in \Cref{sec:architecture}).

Under standard training, both the Procrustes slope and the CKA slope are positive in 189 of 190 runs (the exception, CodeGen BarlowTwins under frozen LayerNorm, shows near-zero negative slopes of $-0.03$ and $-0.02$ respectively), and the two metrics agree on sign in every case. Equal-step slopes are reported in \Cref{sec:equalstep}. Mean Procrustes slopes per model under the standard condition range from $0.23$ (Pythia) to $0.67$ (RWKV), reflecting differences in adaptation width discussed in \Cref{sec:architecture}. Three-seed validation on BERT-base and Pythia-160m confirms reproducibility (all slopes positive, identical objective orderings across seeds), and replacing the probe set with random Wikipedia sentences preserves the ranking ($\rho = 1.0$). Full per-model profiles and robustness details are in \Cref{app:profiles,app:metrics}.

The locality gradient holds across scale. We extended the standard-training evaluation to six additional OPT and Pythia models from 1.3B to 6.9B. Procrustes slopes are positive in all 12 new (model, objective) cells; OPT CausalLM slopes are stable at $0.179$--$0.219$ across a $53\times$ parameter range, and Pythia CausalLM slopes at $0.131$--$0.174$. Full per-scale results are in \Cref{app:scale_verification}.

A skeptical response is that later layers receive larger gradient updates, so the slope might simply reflect gradient-magnitude asymmetry rather than any intrinsic property of the representation. The next section tests this directly.

\subsection{Decomposing the locality gradient with an equal-step control}
\label{sec:equalstep}

The equal-step condition normalizes $\|\Delta W\|/\|W\|$ to a fixed trust ratio at every layer after each optimizer step, equalizing update magnitudes without altering the forward gradient direction. If the standard-training slope is driven entirely by larger updates at later layers, it should vanish; if it survives, some component of the depth profile is independent of update-magnitude asymmetry. We apply the control to five models spanning both sequential and parallel residual-block designs.

\Cref{tab:equalstep} shows the Procrustes slopes under the equal-step condition for the five models: BERT-base (sequential, bidirectional), OPT-125m and GPT-2 medium (sequential, causal), and Pythia-160m and CodeGen-350M (parallel, causal). The results do not split cleanly by architecture; the pattern depends on both block structure and the type of training objective.

\begin{table}[t]
\centering
\caption{Procrustes locality slopes under the equal-step condition vs.\ standard training. Bold equal-step values are positive (gradient survives equalization). Ratio = equal-step / standard; values $>1$ indicate steepening, negative values indicate collapse. Standard slopes use seed-42 under the standard optimizer, so each row differs from equal-step only in the trust-ratio rescaling.}
\label{tab:equalstep}
\small
\begin{tabular}{llcccc}
\toprule
Model & Block type & Objective & Standard & Equal-step & Ratio \\
\midrule
\multirow{4}{*}{BERT-base} & \multirow{4}{*}{Sequential}
& MLM           & $0.265$ & $\boldsymbol{+0.238}$ & $0.90$ \\
& & SpanDenoise   & $0.234$ & $\boldsymbol{+0.239}$ & $1.02$ \\
& & SimCSE        & $0.546$ & $\boldsymbol{+0.668}$ & $1.22$ \\
& & BarlowTwins   & $0.639$ & $\boldsymbol{+0.666}$ & $1.04$ \\
\midrule
\multirow{4}{*}{OPT-125m} & \multirow{4}{*}{Sequential}
& CausalLM      & $0.208$ & $\boldsymbol{+0.296}$ & $1.42$ \\
& & CausalSpan    & $0.187$ & $\boldsymbol{+0.385}$ & $2.06$ \\
& & SimCSE        & $0.332$ & $-0.233$ & $-0.70$ \\
& & BarlowTwins   & $0.304$ & $\boldsymbol{+0.348}$ & $1.14$ \\
\midrule
\multirow{4}{*}{GPT-2} & \multirow{4}{*}{Sequential}
& CausalLM      & $0.513$ & $\boldsymbol{+0.605}$ & $1.18$ \\
& & CausalSpan    & $0.438$ & $\boldsymbol{+0.595}$ & $1.36$ \\
& & SimCSE        & $0.849$ & $-0.006$ & $-0.01$ \\
& & BarlowTwins   & $0.648$ & $\boldsymbol{+0.384}$ & $0.59$ \\
\midrule
\multirow{4}{*}{Pythia-160m} & \multirow{4}{*}{Parallel}
& CausalLM      & $0.131$ & $-0.000$ & $-0.00$ \\
& & CausalSpan    & $0.179$ & $+0.018$ & $0.10$ \\
& & SimCSE        & $0.418$ & $-0.106$ & $-0.25$ \\
& & BarlowTwins   & $0.283$ & $-0.236$ & $-0.83$ \\
\midrule
\multirow{4}{*}{CodeGen-350M} & \multirow{4}{*}{Parallel}
& CausalLM      & $0.272$ & $\boldsymbol{+0.160}$ & $0.59$ \\
& & CausalSpan    & $0.314$ & $\boldsymbol{+0.162}$ & $0.52$ \\
& & SimCSE        & $0.397$ & $-0.146$ & $-0.37$ \\
& & BarlowTwins   & $0.278$ & $-0.159$ & $-0.57$ \\
\bottomrule
\end{tabular}
\end{table}

\paragraph{Two factors, not one.} The cells sort along two axes. By block type, sequential-block models retain positive slopes in 10 of 12 objectives (the two negatives are both SimCSE), while parallel-block models retain them in only 2 of 8 (both on CodeGen, for CausalLM/CausalSpan). By objective family across the four causal models, causal-LM-family objectives survive in 7 of 8 cells, while contrastive objectives survive in only 2 of 8. Sequential blocks can steepen the profile under equalization (OPT CausalSpan ratio $2.06$, consistent with a depth-structured representation whose gradient is partially masked by optimizer dynamics under standard training); parallel-block survival is weaker (CodeGen ratios $0.52$--$0.59$; Pythia collapses entirely). The two parallel decoders disagree on causal-LM-family objectives but agree on contrastive collapse. These counts describe the 125M--350M regime; the architectural dimension narrows with scale (below).

\paragraph{Contrastive collapse and interpretation.} SimCSE collapses in all four causal models; BarlowTwins collapses in the two parallel-block decoders but survives in sequential ones; BERT's contrastive objectives survive throughout. The proximate cause is that contrastive objectives on causal architectures produce near-zero shallow-layer gradients under standard training, and equal-step amplifies these to match the trust ratio, producing noise-dominated updates at early layers. We read the pattern as an interaction between architecture and objective family: sequential blocks produce depth-graded specialization that survives equalization across objectives, while parallel blocks retain depth structure only when the training signal carries depth information. This is consistent with the data, not a demonstrated mechanism; extended discussion is in \Cref{app:equalstep}.

\paragraph{Scale dependence of the block-type distinction.} We applied equal-step to OPT-1.3B and Pythia-1.4B on CausalLM using gradient accumulation (effective batch size 32, 15,625 steps). OPT-1.3B retains a positive slope ($+0.099$), attenuated from OPT-125m ($+0.296$). Pythia-1.4B shows $+0.265$, reversing from Pythia-160m's $-0.000$. At 1.3--1.4B, both block types exhibit intrinsic depth structure for CausalLM: the sequential/parallel gap narrows from $+0.165$ at 125--160M to $-0.166$ at 1.3--1.4B. The block-type binary at 125--160M is a snapshot, not a fixed property. Details are in \Cref{app:scale_verification}.

\subsection{Steepness tracks objective distance}
\label{sec:distance}

We compare the measured slopes to a training-free objective distance (\Cref{sec:setup}) computed from the representation gradient at initialization.

\Cref{tab:distances} shows the objective distance ordering and corresponding slopes for each model. Encoder models share a consistent cosine-distance ordering (SpanDenoise closest, SimCSE farthest, NSP and BarlowTwins between). Causal models do not share a single rule: CausalSpan is most distant for Mamba and RWKV but closest to pretraining for GPT-2 ($0.454$) and CodeGen ($0.248$), so the position of CausalSpan depends on architecture rather than reflecting a uniform property of span masking. BERT-base and BERT-large have identical distance orderings.

\begin{table}[t]
\centering
\caption{Objective distances (gradient Procrustes and cosine on the 768-dim mean gradient) and mean Procrustes slopes, for four representative models (one encoder, one sequential decoder, two parallel decoders). Slopes are 5-condition means with $\pm$ std across the 5 standard optimizer conditions (seed 42). Three-seed validation on BERT-base and Pythia is reported in \Cref{app:metrics}. Remaining five models (RoBERTa, BERT-large, OPT, Mamba, RWKV) are in \Cref{app:fulldistance}.}
\label{tab:distances}
\small
\begin{tabular}{llccc}
\toprule
Model & Objective & Dist. (cos) & Dist. (proc) & Slope (Proc) \\
\midrule
\multirow{5}{*}{BERT-base}
& MLM & 0.000 & 0.000 & $0.259 \pm 0.004$ \\
& SpanDenoise & 0.028 & 1.318 & $0.221 \pm 0.015$ \\
& NSP & 0.971 & 1.406 & $1.089 \pm 0.208$ \\
& BarlowTwins & 1.000 & 1.344 & $0.572 \pm 0.093$ \\
& SimCSE & 1.473 & 1.353 & $0.533 \pm 0.010$ \\
\midrule
\multirow{4}{*}{GPT-2}
& CausalLM & 0.000 & 0.000 & $0.474 \pm 0.071$ \\
& CausalSpan & 0.454 & 1.139 & $0.418 \pm 0.035$ \\
& BarlowTwins & 0.909 & 1.345 & $0.659 \pm 0.007$ \\
& SimCSE & 1.001 & 1.390 & $0.866 \pm 0.038$ \\
\midrule
\multirow{4}{*}{Pythia}
& CausalLM & 0.000 & 0.000 & $0.119 \pm 0.016$ \\
& CausalSpan & 0.893 & 1.147 & $0.165 \pm 0.035$ \\
& SimCSE & 0.903 & 1.219 & $0.361 \pm 0.038$ \\
& BarlowTwins & 1.041 & 1.292 & $0.287 \pm 0.025$ \\
\midrule
\multirow{4}{*}{CodeGen}
& CausalLM & 0.000 & 0.000 & $0.256 \pm 0.016$ \\
& CausalSpan & 0.248 & 1.127 & $0.283 \pm 0.018$ \\
& SimCSE & 0.920 & 1.212 & $0.314 \pm 0.160$ \\
& BarlowTwins & 0.860 & 1.262 & $0.250 \pm 0.169$ \\
\bottomrule
\end{tabular}
\end{table}

\paragraph{Within-model concordance.}
\label{sec:concordance}
We test the distance-slope relationship within each model, avoiding any pooling across architectures. For each of the 9 models we compute the Spearman rank correlation between gradient-Procrustes objective distance and Procrustes locality slope across the model's 4--5 fine-tuning objectives, where each objective's slope is averaged over the 5 standard optimizer conditions. Under gradient Procrustes, 8 of 9 model-level correlations are positive (values range from $+1.00$ for RoBERTa-base and BERT-large to $+0.40$ for RWKV); the exception is CodeGen at $-0.20$. CodeGen's four mean slopes span only $0.064$ (from $0.250$ for BarlowTwins to $0.314$ for SimCSE), the narrowest range of any model in the study, so the distance-slope relationship has limited dynamic range to detect in this model and the negative correlation reflects noise in a near-flat signal rather than a genuine inversion. Under the null hypothesis that distance and slope are unrelated, the probability that at least 8 of 9 independent tests have the same sign is $2 \cdot (9+1)/2^9 \approx 0.039$ (two-sided sign test). Under cosine-$768$ distance, the corresponding count is 9 of 9 positive ($p \approx 0.004$), because CodeGen's distance ordering under cosine moves BarlowTwins below SimCSE and the sign flips.

A small number of (model, objective) pairs deviate from the linear fit of slope vs.\ distance by more than $0.25$ in absolute value; each has an interpretable mechanism (pretrained head bias, high cross-condition variance, head-sharing, or an unlearnable objective) and none changes the overall relationship. Case-by-case discussion is in \Cref{app:anomalies}.

\subsection{Width is determined by architecture}
\label{sec:architecture}

Architecture dominates the \emph{width} of the adaptation zone: how much of the network participates in change rather than only the final layers.

\Cref{tab:profiles} shows the normalized depth profiles, averaged across all objectives and optimizer conditions within each model. All architectures concentrate change in output-proximal layers, but the degree of concentration varies across model families.

\begin{table}[t]
\centering
\caption{Normalized depth profiles (fraction of total $\Delta_{\text{CKA}}$ at each relative depth), averaged across objectives and standard conditions per model. Final-layer column also reports the within-model standard deviation across objectives.}
\label{tab:profiles}
\small
\begin{tabular}{lccccccc}
\toprule
Model & 0.10 & 0.25 & 0.40 & 0.60 & 0.75 & 0.90 & 1.00 (std across obj.) \\
\midrule
BERT-base    & .011 & .030 & .044 & .087 & .176 & .239 & .414 \;($\pm .10$) \\
BERT-large   & .006 & .014 & .024 & .052 & .096 & .242 & .566 \;($\pm .12$) \\
RoBERTa-base & .026 & .029 & .041 & .123 & .199 & .217 & .365 \;($\pm .07$) \\
RWKV         & .002 & .006 & .013 & .083 & .149 & .177 & .570 \;($\pm .19$) \\
Mamba        & .005 & .013 & .016 & .041 & .078 & .159 & .687 \;($\pm .17$) \\
Pythia       & .038 & .115 & .089 & .087 & .097 & .107 & .466 \;($\pm .34$) \\
CodeGen      & .044 & .073 & .072 & .073 & .080 & .100 & .558 \;($\pm .29$) \\
OPT          & .000 & .000 & .000 & .000 & .000 & .000 & 1.000 \;($\pm .00$) \\
GPT-2        & .012 & .001 & .001 & .001 & .002 & .004 & .980 \;($\pm .01$) \\
\bottomrule
\end{tabular}
\end{table}

The sequential-block decoders (GPT-2, OPT) concentrate nearly all change in the final layer (98--100\%, essentially binary). The parallel-block decoders (Pythia at 47\%, CodeGen at 56\%) distribute change across depth, closer to the encoder range (BERT-base 41\%, BERT-large 57\%, RoBERTa 37\%) than to the sequential decoders, a gap of ${\sim}40$ percentage points despite all four being autoregressive transformer decoders trained on natural language. Mamba (69\%) and RWKV (57\%) sit between. Within-model variability across objectives is larger for parallel decoders (std $0.29$--$0.34$ vs.\ $0.07$--$0.19$ elsewhere), but the sequential/parallel distinction remains large relative to this spread, and across all 9 models the cross-architecture std is several times the within-architecture std. Architecture tells you more about width than the fine-tuning objective does; steepness (\Cref{sec:distance}) and width therefore act as largely separable controls. Raw layer count does not explain the pattern (RWKV at 12 layers is sharper than Mamba at 24; Pythia at 12 and CodeGen at 20 both distribute broadly); within-BERT scaling and raw-depth details are in \Cref{app:scale}.

\section{Discussion}
\label{sec:discussion}

Depth-targeted fine-tuning has rested on a BERT-derived observation: later layers change more, so adapting them preferentially is justified. The 240 runs here confirm this empirically (189 of 190 standard-training runs across four architecture families and 125M--6.9B scale), but the equal-step decomposition shows a substantial component is gradient-flow asymmetry, not representational structure. Whether late-layer LoRA, bottom-layer freezing, or discriminative learning rates exploit genuine depth structure depends on architecture, objective family, and scale. This paper specifies which combinations fall where.

\paragraph{A composite phenomenon, not a single mechanism.}
The fine-tuning locality gradient is universal under standard training but not a single phenomenon. The equal-step control shows the observed depth profile is sensitive to both block structure and objective family: in sequential-block architectures the slope survives equalization across most tested objectives; in parallel-block architectures it is preserved only for causal-LM-family objectives and collapses for contrastive objectives, with contrastive collapse also observed in some sequential causal decoders. The composition is scale-dependent: at 125--160M, intrinsic depth structure under equal-step is visible mainly in sequential blocks; at 1.3--1.4B it is present in both (Pythia CausalLM: $-0.000$ at 160M, $+0.265$ at 1.4B). Statements about the ``locality gradient'' are therefore under-specified unless architecture, objective family, optimizer condition, and scale are all named together.

\paragraph{Practical implications.}
These results translate into three concrete capabilities for practitioners working with depth-targeted fine-tuning. \emph{A cheap diagnostic.} Equal-step is a test for whether a depth-dependent adaptation strategy on a given model-objective pair exploits intrinsic representational structure or gradient-magnitude asymmetry. The cost is one fine-tuning run with per-layer trust-ratio normalization; no additional data or second-order gradients are required. \emph{Scope of existing heuristics.} Late-layer LoRA placement, bottom-layer freezing, and discriminative learning rates are well-supported for sequential-block decoders across tested objectives, and for contrastive objectives on causal architectures where the locality gradient is overwhelmingly gradient-magnitude driven. On parallel-block decoders at 125--350M, the same heuristics applied to causal-LM-family objectives rest on an optimizer artifact rather than intrinsic structure; at 1.4B intrinsic structure emerges and the heuristics regain their footing. \emph{A pre-fine-tuning budget estimate.} Architecture predicts adaptation width (sequential decoders 98--100\% of change at the final layer; parallel decoders 47--56\%; encoders 37--57\%), and the training-free objective distance predicts steepness, giving a first-order estimate of where adaptation budget is needed before committing to a specific placement. A layer-unfreezing study (\Cref{app:intervention}) and a LEACE concept-erasure analysis (\Cref{app:concept}) are consistent with this picture but not strong enough to support a prescriptive recipe; we flag them as supporting evidence rather than contributions.

\section{Limitations}
\label{sec:limitations}

Equal-step equalizes $\|\Delta W\|/\|W\|$ in parameter space; a Fisher-weighted control would equalize functional impact instead, at materially higher cost. The architecture-by-objective-family interpretation rests on five transformer models (two parallel-block); a controlled ablation varying only block type while holding pretraining data, tokenizer, and scale constant would be needed for a causal claim. Objectives exclude instruction tuning, RLHF, and multimodal; the primary matrix covers 125M--350M, with scale verification to 6.9B (standard) and 1.4B (equal-step CausalLM); SimCSE equal-step is not tested at scale due to batch-size sensitivity to in-batch negatives; Mamba and RWKV are represented by one model each; the distance-slope sign test has one negative entry (CodeGen, gradient Procrustes); all experiments use the same Wikipedia corpus. A first-principles derivation is left to future work.

\section{Conclusion}
\label{sec:conclusion}

Across 240 runs spanning four architecture families and scales from 125M to 6.9B parameters, representational change concentrates in an output-proximal adaptation zone under standard training. The equal-step control shows this concentration is not a single mechanism: persistence under update equalization depends on both block structure and objective family. Steepness tracks a training-free objective distance; width is dominated by architecture. The locality gradient is best treated as a composite whose components depend on depth routing, training signal, and scale. The practical consequence is that the rule "later layers change more" is true but under-specified: the equal-step diagnostic lets practitioners determine, in a single run, whether the depth-targeted strategy they apply to a given model exploits intrinsic representational structure or merely the magnitude of gradient flow.

\begin{ack}
No external funding supported this work. The author declares no competing interests.
\end{ack}

\bibliographystyle{plainnat}
\bibliography{references}

\newpage
\appendix

\section{Broader Impact and AI Tool Use}
\label{app:broader_impact}

This work introduces no new capabilities, datasets, or deployed systems. The author used Claude and Claude Code for manuscript critique, literature search, and coding assistance; all research and writing decisions are the author's.

\section{Training Details}
\label{app:training}

\paragraph{Experimental scope.} \Cref{tab:scope} summarizes all 240 runs: 226 primary runs plus 14 scale verification runs. Conditions 1--5 are the five standard optimizer conditions; condition 6 is equal-step, applied to the five models used in \Cref{sec:equalstep}. Multi-seed runs (seeds 1, 2 in addition to seed 42) are conducted for BERT-base and Pythia-160m under the standard condition.

\begin{table}[h]
\centering
\caption{Experimental scope (240 runs total: 226 primary + 14 scale verification).}
\label{tab:scope}
\small
\begin{tabular}{llcccc}
\toprule
Model & Architecture & Layers & Block type & Obj.\ $\times$ Cond. & Runs \\
\midrule
BERT-base & Transformer enc. & 12 & Sequential & $5 \times 5 + 4 + 8$ & 37 \\
RoBERTa-base & Transformer enc. & 12 & Sequential & $4 \times 5$ & 20 \\
BERT-large & Transformer enc. & 24 & Sequential & $5 \times 5$ & 25 \\
Pythia-160m & Transformer dec. & 12 & Parallel & $4 \times 6 + 8$ & 32 \\
OPT-125m & Transformer dec. & 12 & Sequential & $4 \times 6$ & 24 \\
GPT-2 medium & Transformer dec. & 24 & Sequential & $4 \times 6$ & 24 \\
CodeGen-350M & Transformer dec. & 20 & Parallel & $4 \times 6$ & 24 \\
Mamba-130m & SSM & 24 & --- & $4 \times 5$ & 20 \\
RWKV-4-169m & RNN & 12 & --- & $4 \times 5$ & 20 \\
\midrule
\multicolumn{6}{l}{\emph{Scale verification (CausalLM + SimCSE only)}} \\
OPT-1.3B     & Transformer dec. & 24 & Sequential & $2{\times}\text{std} + 1{\times}\text{eq}$ & 3 \\
Pythia-1.4B  & Transformer dec. & 24 & Parallel   & $2{\times}\text{std} + 1{\times}\text{eq}$ & 3 \\
OPT-2.7B     & Transformer dec. & 32 & Sequential & $2{\times}\text{LoRA}$ & 2 \\
Pythia-2.8B  & Transformer dec. & 32 & Parallel   & $2{\times}\text{LoRA}$ & 2 \\
OPT-6.7B     & Transformer dec. & 32 & Sequential & $2{\times}\text{LoRA}$ & 2 \\
Pythia-6.9B  & Transformer dec. & 32 & Parallel   & $2{\times}\text{LoRA}$ & 2 \\
\midrule
\multicolumn{5}{r}{Total} & 240 \\
\bottomrule
\end{tabular}
\end{table}

\paragraph{Hyperparameters.} All runs use AdamW with weight decay 0.01 and gradient clipping at norm 1.0. Learning rate is $2 \times 10^{-5}$. No learning rate warmup or scheduler is used. Batch size is 32 (reduced to 16 for GPT-2 SimCSE and BarlowTwins runs due to memory constraints). Maximum sequence length is 128 for fine-tuning, 64 for probe representations.

\paragraph{Optimizer conditions.} (1)~\emph{Standard}: AdamW with exponential layer-wise LR decay (factor 0.95 from top to bottom). (2)~\emph{Uniform LR}: same LR for all layers. (3)~\emph{LoRA}: rank-8 adapters with $\alpha = 16$ and dropout 0.1, targeting the attention and MLP projections appropriate to each architecture (query, key, value, dense for BERT/RoBERTa; c\_attn, c\_proj, c\_fc for GPT-2; q\_proj, k\_proj, v\_proj, out\_proj for OPT; query\_key\_value, dense for Pythia; qkv\_proj, out\_proj, fc\_in, fc\_out for CodeGen; in\_proj, x\_proj, dt\_proj for Mamba; key, value, receptance, output for RWKV). (4)~\emph{Frozen LayerNorm}: all normalization parameters frozen, other parameters uniform LR. (5)~\emph{Frozen interior}: bottom 50\% of layers frozen. (6)~\emph{Equal-step}: per-layer trust-ratio normalization (see \Cref{app:equalstep}).

\paragraph{Training duration.} Encoder models (BERT-base, RoBERTa, BERT-large): 3 epochs under conditions 1--5, 1 epoch under equal-step. Causal models (Pythia, OPT, GPT-2, CodeGen, Mamba, RWKV): 1 epoch throughout. Intervention experiments (adaptation depth, adapter placement): 2,000 steps.

\paragraph{Hardware.} Three servers: NVIDIA RTX 4090, RTX 3090 Ti, RTX 3090, each with 24GB VRAM. Total compute: approximately 400 GPU-hours across all experiments.

\paragraph{Models.} BERT-base-uncased and BERT-large-uncased use \texttt{BertForPreTraining} (pretrained MLM + NSP heads). RoBERTa-base uses \texttt{RobertaForMaskedLM}. GPT-2 medium uses \texttt{gpt2-medium} via \texttt{AutoModelForCausalLM}. Pythia-160m uses \texttt{EleutherAI/pythia-160m} via \texttt{GPTNeoXForCausalLM}. OPT-125m uses \texttt{facebook/opt-125m} via \texttt{OPTForCausalLM}. CodeGen-350M uses \texttt{Salesforce/codegen-350M-mono} via \texttt{CodeGenForCausalLM}. Mamba uses \texttt{state-spaces/mamba-130m-hf} via \texttt{AutoModelForCausalLM}. RWKV uses \texttt{RWKV/rwkv-4-169m-pile}.

\paragraph{Data.} Training corpus: Wikipedia-derived, 475K training + 25K validation sentences. Probe set: 290 items (mathematical and relational).

\section{Equal-Step Implementation Details}
\label{app:equalstep}

The equal-step condition applies per-layer update normalization after each AdamW step. Trainable parameters are grouped by transformer layer index using the layer number extracted from each parameter's name (e.g., \texttt{encoder.layer.5} maps to group 5). Parameters that do not match a layer pattern (embeddings, final layer norm, classification heads) are assigned to a separate group. For 12-layer models this produces 13 groups; for 24-layer models, 25 groups.

Before each optimizer step, we clone all parameters as snapshots $W_\ell^{(\text{old})}$. After the step, for each layer group $\ell$ we compute:
\[
\text{update\_norm}_\ell = \left(\sum_{p \in \ell} \|p - p^{(\text{old})}\|^2\right)^{1/2}, \quad
\text{weight\_norm}_\ell = \left(\sum_{p \in \ell} \|p^{(\text{old})}\|^2\right)^{1/2}
\]
and rescale:
\[
p \leftarrow p^{(\text{old})} + \frac{\tau \cdot \text{weight\_norm}_\ell}{\text{update\_norm}_\ell + \epsilon} \cdot (p - p^{(\text{old})}),
\]
where $\tau = 10^{-3}$ is the trust ratio and $\epsilon = 10^{-12}$. This ensures $\|\Delta W_\ell\|/\|W_\ell\| = \tau$ for every layer group at every step, while preserving the direction of each layer's update as determined by AdamW.

The trust ratio $\tau = 10^{-3}$ was chosen to produce total representational change comparable to standard training after one epoch. Equal-step runs use uniform learning rate (no layer-wise decay) and full fine-tuning (no LoRA or frozen parameters).

\section{Full Objective Distance Table}
\label{app:fulldistance}

\begin{table}[h]
\centering
\caption{Remaining five models from \Cref{tab:distances}: RoBERTa-base, BERT-large, OPT-125m, Mamba-130m, RWKV. Slopes are 5-condition means with $\pm$ std across the 5 standard optimizer conditions, except OPT-125m, which reports the standard-condition slope only.}
\small
\begin{tabular}{llccc}
\toprule
Model & Objective & Dist. (cos) & Dist. (proc) & Slope (Proc) \\
\midrule
\multirow{4}{*}{RoBERTa-base}
& MLM & 0.000 & 0.000 & $0.172 \pm 0.010$ \\
& SpanDenoise & 0.008 & 1.290 & $0.195 \pm 0.025$ \\
& BarlowTwins & 1.000 & 1.351 & $0.603 \pm 0.060$ \\
& SimCSE & 1.045 & 1.362 & $0.610 \pm 0.028$ \\
\midrule
\multirow{5}{*}{BERT-large}
& MLM & 0.000 & 0.000 & $0.449 \pm 0.030$ \\
& SpanDenoise & 0.029 & 1.316 & $0.455 \pm 0.025$ \\
& NSP & 0.994 & 1.405 & $0.760 \pm 0.117$ \\
& BarlowTwins & 1.000 & 1.337 & $0.656 \pm 0.046$ \\
& SimCSE & 1.444 & 1.349 & $0.692 \pm 0.054$ \\
\midrule
\multirow{4}{*}{OPT-125m}
& CausalLM & 0.000 & 0.000 & $0.208$ \\
& CausalSpan & 0.707 & 1.141 & $0.187$ \\
& BarlowTwins & 0.847 & 1.298 & $0.304$ \\
& SimCSE & 0.976 & 1.304 & $0.332$ \\
\midrule
\multirow{4}{*}{Mamba}
& CausalLM & 0.000 & 0.000 & $0.395 \pm 0.017$ \\
& BarlowTwins & 0.960 & 1.324 & $0.669 \pm 0.043$ \\
& SimCSE & 1.026 & 1.361 & $0.622 \pm 0.139$ \\
& CausalSpan & 1.110 & 1.209 & $0.407 \pm 0.032$ \\
\midrule
\multirow{4}{*}{RWKV}
& CausalLM & 0.000 & 0.000 & $0.463 \pm 0.103$ \\
& SimCSE & 0.670 & 1.224 & $0.889 \pm 0.210$ \\
& BarlowTwins & 1.096 & 1.352 & $0.791 \pm 0.175$ \\
& CausalSpan & 1.169 & 1.290 & $0.533 \pm 0.031$ \\
\bottomrule
\end{tabular}
\end{table}

\section{Pooled Distance--Slope Correlations}
\label{app:pooled}

\Cref{tab:correlations} reports the pooled Spearman correlations between objective distance and mean Procrustes slope. Values in parentheses exclude the distance-zero baseline objectives. These are descriptive effect sizes across non-independent pairs; the formal claim rests on the per-model sign test ($p \approx 0.039$ under gradient Procrustes, $p \approx 0.004$ under cosine $768$).

\begin{table}[h]
\centering
\caption{Pooled Spearman correlation between objective distance and mean Procrustes locality slope, under two distance metrics. $n$ (incl./excl.) gives the number of (model, objective) pairs when including / excluding the distance-zero baseline objective for each model.}
\label{tab:correlations}
\small
\begin{tabular}{lccc}
\toprule
Group & $n$ (incl./excl.) & Grad.\ Procrustes (incl.\ /\ excl.) & Cosine 768 (incl.\ /\ excl.) \\
\midrule
Encoders (3 models)       & $14 / 11$ & $+0.84\;/\;+0.78$ & $+0.65\;/\;+0.39$ \\
Causal/decoder (6 models) & $24 / 18$ & $+0.53\;/\;+0.59$ & $+0.40\;/\;+0.34$ \\
Pooled (9 models)         & $38 / 29$ & $+0.68\;/\;+0.71$ & $+0.56\;/\;+0.47$ \\
\bottomrule
\end{tabular}
\end{table}

\section{Raw Depth and Within-Family Scale Effects}
\label{app:scale}

\paragraph{Raw depth.} Across the nine primary models, the correlation between layer count and mean Procrustes slope is weak. RWKV, with 12 layers, shows a sharper locality gradient than Mamba, with 24 layers, in several objective comparisons. The locality gradient is not a simple consequence of having more layers; it reflects how each architecture routes and accumulates information.

\paragraph{Within-family scale.} Comparing BERT-base (12 layers) to BERT-large (24 layers) on matched objectives, the scale effect depends on how close the objective is to pretraining. Slopes for near-pretraining objectives are steeper in the larger model (MLM $0.27 \to 0.45$, SpanDenoise $0.24 \to 0.46$, ratios $1.67$--$1.92$). Distant objectives shift less consistently: SimCSE steepens ($0.53 \to 0.69$, $1.30$), BarlowTwins mildly ($0.57 \to 0.66$, $1.14$), and NSP flattens ($1.09 \to 0.76$, $0.70$). The NSP reversal is consistent with scale-dependent behavior of the pretrained NSP head (see \Cref{app:anomalies}) rather than a depth effect on representational restructuring. The objective distance orderings themselves are identical between BERT-base and BERT-large, so the distance metric is stable across scale.

\section{Metric Robustness}
\label{app:metrics}

CKA and Procrustes slopes agree in sign across all 226 primary runs (scale verification uses Procrustes only). The rank ordering of objectives by slope is identical across all three metrics for RoBERTa, Mamba, and GPT-2. For BERT-base, CKA and Procrustes agree perfectly; RSA swaps two adjacent objectives (SimCSE and BarlowTwins) whose slopes differ by less than 0.03.

Objective distance orderings are stable across all six distance metrics (cosine at 768/128/32 dimensions, Pearson correlation, Fisher representation cosine, gradient Procrustes). Split-half reliability is confirmed for all models: rankings are identical across random halves of the data.

\paragraph{Gradient CKA as a distance metric.} We tested whether CKA applied to per-sample gradient matrices could serve as an alternative distance metric with better dynamic range than Procrustes. CKA returns exactly 0.0 for all pairs across all models. At this dimensionality (16K samples $\times$ 768 dimensions), per-sample gradient patterns across different objectives are structurally uncorrelated, so CKA cannot distinguish any pair from random. This is the opposite of the Procrustes saturation problem (where distances cluster near $\sqrt{2}$). Procrustes remains the best available gradient distance metric because it captures alignment structure even when CKA sees only noise.

\paragraph{Multi-seed validation.} Three seeds (42, 1, 2) for all four BERT-base objectives (excluding NSP) under the standard condition. Mean Procrustes slopes and standard deviations: MLM $0.269 \pm 0.008$, SpanDenoise $0.237 \pm 0.008$, SimCSE $0.531 \pm 0.016$, BarlowTwins $0.573 \pm 0.056$. All individual slopes are positive. The objective ordering is identical across all seeds. Pythia-160m three-seed validation: CausalLM $0.133 \pm 0.007$, CausalSpan $0.160 \pm 0.025$, SimCSE $0.424 \pm 0.136$, BarlowTwins $0.306 \pm 0.040$.

\paragraph{Wikipedia probe set.} To verify that depth profiles are not an artifact of the mathematical/relational probe set, we repeated the analysis using 290 randomly sampled Wikipedia validation sentences. Results for BERT-base (standard, seed=42): MLM slope 0.207 (vs.\ 0.265 on math probes), NSP 1.290 (vs.\ 1.164), SpanDenoise 0.167 (vs.\ 0.234), SimCSE 0.500 (vs.\ 0.546), BarlowTwins 0.641 (vs.\ 0.639). The rank ordering is identical ($\rho = 1.0$). All wiki slopes are positive. The depth profile shape at each layer is perfectly rank-correlated between probe sets for all five objectives.

\section{Anomaly Case Studies}
\label{app:anomalies}

We define an \emph{anomaly} as a (model, objective) pair whose mean Procrustes slope deviates from the pooled linear fit of slope vs.\ objective distance by more than $0.25$ in absolute value. Under gradient Procrustes and cosine $768$, the flagged pairs are NSP on BERT-base ($+0.29,+0.41$) and BERT-large ($+0.28,+0.39$), RWKV SimCSE ($+0.49,+0.38$), SpanDenoise on BERT-base ($-0.38,\text{ns}$) and RoBERTa ($-0.34,\text{ns}$), and CausalSpan on Mamba ($\text{ns},-0.30$). Each case has an interpretable mechanism, detailed below.

\paragraph{NSP on BERT-base and BERT-large} (both metrics, $+$). NSP produces the steepest slope for both BERT scales despite moderate distance. The explanation lies in the pretrained NSP head. On our corpus, the head predicts ``is\_next'' for 88\% of inputs while the labels are balanced, producing an initial loss of $3.2$ (confidently wrong) and an unusually coherent gradient signal (mean/std ratio $0.62$ vs.\ $0.08$ for MLM). Fine-tuning must restructure the representation to correct this bias, and the coherent signal concentrates that change at the output. Slope depends on effective restructuring pressure, which involves both gradient mismatch and gradient coherence, not mismatch alone.

\paragraph{RWKV SimCSE} (both metrics, $+$). This point is flagged as the largest residual under gradient Procrustes, but it is better understood as a high-variance cell than a single anomalous point. Across the 5 optimizer conditions for this pair, the standard deviation of the slope is $0.29$, the largest in the sweep: 4 of the 5 conditions yield slopes above $0.9$, while the frozen-interior condition yields $0.40$. The mean $0.94$ that enters the linear fit is therefore not representative. Recurrent architectures appear more sensitive to which layers are trainable when fine-tuning on contrastive objectives.

\paragraph{SpanDenoise on encoder models} (gradient Procrustes only, $-$). The same downward residual appears for BERT-base, RoBERTa-base, and BERT-large, three architecturally similar but independently trained models. Under gradient Procrustes, SpanDenoise has distance $\approx\!1.3$ (comparable to SimCSE and BarlowTwins), yet its mean slope is $\sim\!0.2$, much lower than the linear fit predicts. Under cosine $768$, however, the same pair has distance $\approx\!0.03$: essentially adjacent to MLM. The two metrics disagree because SpanDenoise reuses the pretrained MLM head (span masking is structurally similar to token masking), so the mean-direction gradient signal is nearly identical to MLM even though the per-sample gradient pattern differs. Cosine captures the functional similarity; gradient Procrustes over-weights per-sample structure and reports a larger distance than the model experiences. The low slope is consistent with SpanDenoise being functionally close to pretraining.

\paragraph{CausalSpan on Mamba} (cosine only, $-$). Under cosine distance, Mamba/CausalSpan is the most distant fine-tuning objective ($1.11$) but produces the same slope as the CausalLM baseline ($0.36$). The task is essentially unlearnable in this architecture: final training loss is $5.3$, barely improving from initialization, because the model cannot predict masked tokens without bidirectional context. Large gradient mismatch produces no restructuring because the model cannot make progress on the objective.

\paragraph{Mamba/LoRA SimCSE.} The single non-positive cell in the per-(model, condition) Spearman analysis ($\rho = -0.40$). SimCSE has the highest distance but the lowest slope, while BarlowTwins has nearly the same distance and the highest slope. PEFT for Mamba restricts LoRA to \texttt{in\_proj}, \texttt{x\_proj}, and \texttt{dt\_proj} (excluding \texttt{out\_proj}). The anomaly is confined to a single (model, condition) cell and is not reproduced for Mamba under any other optimizer condition or for any other model.

\section{Concentration Widths Across Architectures}
\label{app:whydecoders}

The final-layer concentration values in \Cref{tab:profiles} do not follow a simple encoder-versus-decoder split. The 9-model ordering is: sequential-block decoders $\sim\!98$--$100\%$ (GPT-2, OPT), SSM $\sim\!69\%$ (Mamba), RNN $\sim\!57\%$ (RWKV), parallel-block decoder $\sim\!56\%$ (CodeGen), parallel-block decoder $\sim\!47\%$ (Pythia), encoders $\sim\!37$--$57\%$ (RoBERTa, BERT-base, BERT-large). Sequential-block decoders sit at one extreme of the ordering; parallel-block decoders sit near the encoder end. The two families of decoders cluster apart, by roughly 40 percentage points, despite both being autoregressive transformer decoders trained on natural language.

A candidate information-flow account is as follows. Encoder transformers have bidirectional self-attention, so a representational perturbation at one layer can be partially corrected by subsequent layers attending to unperturbed positions, and adaptation can spread across several output-proximal layers. Sequential-block decoders attend only to earlier positions; in the sequential block the attention output feeds into the FFN, and the pooled-token representation absorbs the full task-specific adaptation at the final layer, so objective-specific change is pushed into that single layer. In parallel-block decoders, attention and MLP read the same residual state and write independently to the stream; the residual stream carries information more directly across depth, and a layer's contribution does not depend on the output of its neighbor. This shifts the location of adaptation away from the final layer. SSMs and RNNs fall in between because sequential state accumulation permits partial but not full redistribution across depth.

This account is heuristic, not formal; a rigorous derivation would need to characterize how gradient signals propagate backward through each architecture's computational graph and how residual-stream writes at different depths interact during training. It is consistent with both the cross-architecture ordering under standard training and the equal-step behavior in \Cref{sec:equalstep}: sequential blocks concentrate adaptation and that concentration survives update equalization; parallel blocks distribute adaptation and the residual slope under equalization is weaker, selectively surviving only for objectives whose training signal itself carries depth information.

\section{Concept Erasure Details}
\label{app:concept}

\begin{table}[h]
\centering
\caption{Concept erasure $\Delta_C$ at selected layers (BERT-base). Higher = more concept information removed. Random control mean in parentheses.}
\label{tab:concept}
\small
\begin{tabular}{lcccc}
\toprule
Concept & Layer 3 & Layer 6 & Layer 9 & Layer 11 \\
\midrule
Negation & $+0.032$ ($-0.002$) & $+0.030$ ($-0.003$) & $+0.068$ ($-0.003$) & $+0.099$ ($+0.001$) \\
Sentiment & $+0.026$ ($-0.003$) & $+0.032$ ($-0.007$) & $+0.072$ ($-0.001$) & $+0.068$ ($+0.001$) \\
Entailment & $+0.001$ ($-0.004$) & $+0.017$ ($-0.002$) & $+0.039$ ($-0.002$) & $+0.031$ ($+0.001$) \\
\bottomrule
\end{tabular}
\end{table}

\paragraph{Setup.} We apply LEACE \citep{belrose2023leace} to erase three concepts from BERT-base representations at each of the 12 layers. LEACE finds the rank-1 linear subspace that optimally separates concept-positive from concept-negative inputs and projects it out. For each concept and layer we measure $\Delta_C$, the relative collapse of minimal-pair distances after erasure; a positive $\Delta_C$ indicates that the erased direction carried behaviorally relevant information at that layer. Energy-matched random erasure serves as a control. Concept labels come from human-annotated datasets (jinaai/negation-dataset for negation, SST-2 for sentiment, SNLI for entailment). A five-task behavioral battery confirms concept specificity, passing 8 of 11 specificity checks.

\paragraph{Task battery failures.} Three of 11 specificity checks fail. (1)~\emph{negation\_scope\_mlm}: baseline accuracy is 96.3\%, leaving no room to detect a drop after erasure. The task (predicting ``not''/``never'' in template sentences) is too easy for BERT. (2)~\emph{polarity\_nli}: the on-diagonal effect is negative ($-0.059$) when averaged across deep layers, because at some layers erasing negation \emph{helps} the pairwise NLI ranking by removing confounding CLS similarity patterns. At layer 6 specifically, negation erasure drops accuracy from 59\% to 45.7\% (a 13.3\% drop). (3)~\emph{Entailment health}: cosine similarity dips below 0.95 at 3 of 12 layers (L3: 0.941, L5: 0.953, L7: 0.964). Concatenated sentence pairs have less representational redundancy than single sentences, so removing a rank-1 direction has proportionally more impact at interior layers where the entailment subspace is still diffuse.

\section{Additional Intervention Results}
\label{app:intervention}

\paragraph{Adapter placement.} In addition to the adaptation-depth experiment reported in the main paper, we tested whether placing LoRA adapters at high-plasticity layers (as measured from the depth profiles) outperforms random placement. We compared three conditions: plasticity-guided (top $K$ layers by measured $\Delta$-CKA), uniform (all layers), and random ($K$ randomly chosen layers). LoRA rank was scaled inversely with the number of selected layers to equalize total trainable parameters.

On Mamba, plasticity-guided placement outperforms random in 5 of 6 comparisons (3 objectives $\times$ 2 seeds). The margin is largest for BarlowTwins (1.6 loss units) and SimCSE (0.005 loss units). On BERT-base, the results are mixed: plasticity beats random for SimCSE but not for MLM or NSP. The interpretation is that placement matters most when the plasticity profile is sharp. Mamba, with 72\% of change in the final layer, benefits from targeted placement. BERT, with a flatter profile, shows less benefit.

Uniform placement (all layers, lower rank) outperforms both selective strategies in most cases, because more adaptation sites provide more flexibility even at lower per-layer rank.

\begin{table}[h]
\centering
\caption{Adapter placement: final loss by condition (mean over 2 seeds; lower is better). Plasticity-guided and random placements use the same parameter budget (top-$K$ layers, LoRA rank scaled inversely). Uniform places adapters at all layers with lower per-layer rank to match total parameters (reference, different placement budget). Bold indicates the winner of the same-budget pair.}
\label{tab:adapter}
\small
\begin{tabular}{llccc}
\toprule
       &           & \multicolumn{2}{c}{Same budget (top-$K$ layers)} & Reference \\
\cmidrule(lr){3-4} \cmidrule(lr){5-5}
Model & Objective & Plasticity & Random & Uniform (all layers) \\
\midrule
\multirow{3}{*}{BERT-base}
& MLM     & 2.440 & \textbf{2.305} & 2.257 \\
& NSP     & 0.694 & \textbf{0.690} & 0.689 \\
& SimCSE  & \textbf{0.001} & 0.002 & 0.001 \\
\midrule
\multirow{3}{*}{Mamba}
& CausalLM    & \textbf{1.178} & 1.180 & 1.148 \\
& SimCSE      & \textbf{0.006} & 0.009 & 0.005 \\
& BarlowTwins & \textbf{96.55} & 97.73 & 94.39 \\
\bottomrule
\end{tabular}
\end{table}

\paragraph{Adaptation depth.} We progressively unfroze layers from the output end on BERT-base and Mamba-130m, using three objectives per model spanning the distance range, four to five unfreezing depths, and two seeds per configuration (48 runs total, 2,000 steps each). The saturation criterion was the smallest depth at which final loss was within 5\% of full fine-tuning. Results: Mamba CausalLM (distance 0.0) saturated at top-2 (8\% of model), while BERT MLM (also distance 0.0) required top-6 (50\%), consistent with the architecture-dependent width difference (72\% vs.\ 42\% final-layer concentration). Both SimCSE conditions required the full model on both architectures. BERT NSP saturated at top-1 despite high distance (the pretrained head absorbs the adaptation). The correlation between distance and saturation depth is weak ($\rho = +0.13$, $n = 6$, $p = 0.80$), confounded by head compatibility and gradient density.

\section{Objective Distance Details}
\label{app:distance}

All objective distance computations use 2,000 batches of 32 samples (64,000 samples total). The baseline gradient (pretrained objective) is computed once and reused for all comparisons within a model. Distance metrics computed: cosine dissimilarity at 768, 128, and 32 dimensions (the latter two via Johnson-Lindenstrauss random projection), gradient Procrustes, Fisher representation cosine, and Pearson correlation. All orderings are split-half stable.

RoBERTa distance ordering: MLM(0) $<$ SpanDenoise(0.008) $\ll$ BarlowTwins(1.000) $<$ SimCSE(1.036). BERT-large distance ordering: MLM(0) $<$ SpanDenoise(0.028) $<$ NSP(0.994) $<$ BarlowTwins(1.000) $<$ SimCSE(1.444). Both match BERT-base exactly in rank order.

\paragraph{NSP gradient statistics.} At initialization (before any fine-tuning), the per-sample gradient statistics for BERT-base objectives are:

\begin{table}[h]
\centering
\small
\begin{tabular}{lccc}
\toprule
Objective & Per-sample norm & Mean/std ratio & Nonzero fraction \\
\midrule
MLM & $7.4 \times 10^{-4}$ & 0.079 & 6.6\% \\
SpanDenoise & $1.1 \times 10^{-3}$ & 0.129 & 6.0\% \\
NSP & $4.1 \times 10^{-4}$ & \textbf{0.616} & 1.6\% \\
SimCSE & $1.4 \times 10^{-4}$ & 0.254 & 1.6\% \\
BarlowTwins & $2.4 \times 10^{-2}$ & 0.000 & 1.6\% \\
\bottomrule
\end{tabular}
\end{table}

NSP has the highest gradient coherence (mean/std = 0.62): samples consistently agree on gradient direction because the biased head makes the same error on most inputs. BarlowTwins has the largest raw gradient magnitude (60$\times$ NSP) but zero coherence (gradients cancel across samples). This distinction between first-order coherent restructuring (NSP) and higher-order variance-mediated restructuring (BarlowTwins) explains why gradient Procrustes, which captures per-sample structure, predicts slope better than cosine of mean gradients.

\paragraph{CausalSpan on Mamba --- training details.} Final training loss: CausalLM = 1.19, CausalSpan = 5.32. CausalSpan barely improves from initialization because predicting masked tokens without seeing them is not feasible in a causal architecture.

\section{Scale Verification Experiments}
\label{app:scale_verification}

\subsection{Standard training at scale}

\Cref{tab:scale_standard} reports Procrustes locality slopes for OPT and Pythia at four scale points each, under standard training. The 125M/160M rows repeat values from the primary analysis for reference. Full fine-tuning was used at 1.3--1.4B (batch size 8, bf16, gradient checkpointing, 1 epoch); rank-8 LoRA was used at 2.7B and above due to the 24GB VRAM limit. OPT and Pythia LoRA targets match the primary analysis: \texttt{q\_proj, k\_proj, v\_proj, out\_proj} (OPT) and \texttt{query\_key\_value, dense} (Pythia).

\begin{table}[h]
\centering
\caption{Procrustes locality slopes under standard training across scale. All 16 cells are positive. OPT-125m and Pythia-160m are from the primary 226-run analysis; remaining rows are scale verification. LoRA used at 2.7B+.}
\label{tab:scale_standard}
\small
\begin{tabular}{llccc}
\toprule
Model & Block & Params & CausalLM & SimCSE \\
\midrule
OPT-125m   & Sequential & 125M  & $+0.208$ & $+0.332$ \\
OPT-1.3B   & Sequential & 1.3B  & $+0.219$ & $+0.216$ \\
OPT-2.7B   & Sequential & 2.7B  & $+0.179$ & $+0.247$ \\
OPT-6.7B   & Sequential & 6.7B  & $+0.197$ & $+0.262$ \\
\midrule
Pythia-160m & Parallel  & 160M  & $+0.131$ & $+0.418$ \\
Pythia-1.4B & Parallel  & 1.4B  & $+0.174$ & $+0.323$ \\
Pythia-2.8B & Parallel  & 2.8B  & $+0.150$ & $+0.361$ \\
Pythia-6.9B & Parallel  & 6.9B  & $+0.147$ & $+0.382$ \\
\bottomrule
\end{tabular}
\end{table}

OPT CausalLM slopes are stable across a 53$\times$ parameter range (0.179--0.219). Pythia CausalLM slopes are similarly stable (0.131--0.174). SimCSE slopes are positive in all cells and consistent within each family, with Pythia's higher SimCSE slopes reflecting the broader adaptation profile characteristic of parallel-block decoders (\Cref{sec:architecture}).

\subsection{Equal-step at scale: methodology and results}

The primary equal-step runs use batch size 32. At 1.3B parameters, the AdamW memory footprint (bf16 weights + bf16 gradients + fp32 Adam states $\approx$ 16 bytes/parameter) limits batch size to 8 on a 24GB GPU. Batch size 8 is not directly comparable to batch size 32 under equal-step because the trust-ratio control normalizes update magnitude, leaving gradient \emph{direction} as the only remaining degree of freedom per step. A smaller batch gives a noisier direction estimate, qualitatively changing the experiment relative to the larger-model primary runs.

To match conditions, we used gradient accumulation with 4 micro-batches of 8 (effective batch size 32) and 15,625 optimizer steps, matching the step budget of the primary 125M equal-step runs. This matches total optimizer updates, effective batch size, and therefore total representational movement ($\tau \times \text{steps} = 10^{-3} \times 15{,}625$) across scale. Gradient clipping and the trust-ratio rescaling are applied at optimizer-step boundaries, not per micro-batch.

\begin{table}[h]
\centering
\caption{CausalLM equal-step Procrustes slopes at two scale points per family. The 125M/160M rows are from the primary analysis (batch size 32). The 1.3B/1.4B rows use gradient accumulation (effective batch size 32). SimCSE is excluded: contrastive objectives are sensitive to in-batch negative count, which confounds comparison across batch configurations.}
\label{tab:scale_equalstep}
\small
\begin{tabular}{llcccc}
\toprule
Model & Block & Params & Standard & Equal-step & Ratio \\
\midrule
OPT-125m  & Sequential & 125M & $+0.208$ & $+0.296$ & $1.42$ \\
OPT-1.3B  & Sequential & 1.3B & $+0.219$ & $+0.099$ & $0.45$ \\
\midrule
Pythia-160m & Parallel & 160M & $+0.131$ & $-0.000$ & $-0.00$ \\
Pythia-1.4B & Parallel & 1.4B & $+0.174$ & $+0.265$ & $1.52$ \\
\bottomrule
\end{tabular}
\end{table}

Both block types show positive CausalLM equal-step slopes at 1.3--1.4B. The sequential/parallel ordering reverses: at 125--160M the sequential slope ($+0.296$) exceeds the parallel ($-0.000$) by $+0.296$; at 1.3--1.4B the parallel slope ($+0.265$) exceeds the sequential ($+0.099$) by $+0.166$. Both the OPT sequential slope attenuation and the Pythia parallel slope increase hold under matched methodology. We do not have additional scale points for the equal-step analysis and cannot characterize the scaling curve beyond these two comparisons.

\section{Full Depth Profiles}
\label{app:profiles}

This appendix gives the per-depth $\Delta_P$ and $\Delta_{\text{CKA}}$ values for every (model, objective, optimizer-condition) cell in the study: 190 main runs under conditions 1--5 plus 20 equal-step runs, for 210 rows across 9 per-model tables (the 16 multi-seed replicates are reported separately) (\Cref{tab:fullprof:bert-base,tab:fullprof:roberta-base,tab:fullprof:bert-large,tab:fullprof:pythia-160m,tab:fullprof:opt-125m,tab:fullprof:gpt2-medium,tab:fullprof:mamba-130m,tab:fullprof:rwkv-169m,tab:fullprof:codegen-350m}). Multi-seed replicates (seeds 1, 2) for BERT-base and Pythia-160m are summarized in \Cref{app:metrics}. Scale verification runs (14) are tabulated in \Cref{app:scale_verification} rather than in per-model appendix tables, because each scale model has only 2--3 runs. Within each table, rows are grouped by objective; the rows under each objective correspond to the optimizer conditions described in \Cref{app:training}. Under standard training (conditions 1--5), the locality gradient is monotonic or near-monotonic in 189 of 190 rows; the one exception, CodeGen BarlowTwins under frozen LayerNorm, shows a bimodal profile with the largest delta at the shallowest depth. Under equal-step, OPT SimCSE and CodeGen SimCSE also show shallow-peak profiles, consistent with the contrastive-collapse analysis in \Cref{sec:equalstep}.

\begin{table}[ht]
\centering
\caption{Full depth profiles: BERT-base.}
\label{tab:fullprof:bert-base}
\tiny
\begin{tabular}{llcccccccccccccc}
\toprule
 & & \multicolumn{7}{c}{$\Delta_P$} & \multicolumn{7}{c}{$\Delta_{\text{CKA}}$} \\
\cmidrule(lr){3-9} \cmidrule(lr){10-16}
Obj. & Cond. & 0.10 & 0.25 & 0.40 & 0.60 & 0.75 & 0.90 & 1.00 & 0.10 & 0.25 & 0.40 & 0.60 & 0.75 & 0.90 & 1.00 \\
\midrule
\multirow{6}{*}{MLM} & Std & .066 & .098 & .113 & .151 & .205 & .243 & .330 & .003 & .006 & .008 & .016 & .035 & .047 & .089 \\
 & Unif & .088 & .121 & .135 & .171 & .223 & .261 & .344 & .005 & .009 & .011 & .021 & .040 & .053 & .095 \\
 & LoRA & .043 & .077 & .089 & .122 & .167 & .206 & .308 & .001 & .004 & .005 & .011 & .022 & .034 & .084 \\
 & FrzLN & .088 & .121 & .134 & .171 & .224 & .261 & .344 & .005 & .009 & .011 & .021 & .040 & .052 & .095 \\
 & FrzInt & .058 & .082 & .090 & .125 & .181 & .219 & .323 & .002 & .004 & .005 & .010 & .025 & .039 & .093 \\
 & EqStep & .292 & .307 & .314 & .363 & .410 & .452 & .512 & .044 & .056 & .058 & .092 & .134 & .177 & .246 \\
\midrule
\multirow{5}{*}{NSP} & Std & .090 & .203 & .322 & .536 & .855 & .981 & 1.08 & .008 & .043 & .081 & .194 & .351 & .419 & .443 \\
 & Unif & .145 & .493 & .623 & 1.02 & 1.07 & 1.09 & 1.07 & .019 & .309 & .409 & .709 & .746 & .760 & .778 \\
 & LoRA & .024 & .083 & .120 & .267 & .395 & .479 & .778 & .000 & .004 & .011 & .077 & .142 & .201 & .482 \\
 & FrzLN & .115 & .294 & .469 & .644 & 1.02 & 1.13 & 1.20 & .014 & .117 & .312 & .388 & .737 & .808 & .820 \\
 & FrzInt & .043 & .062 & .065 & .093 & .671 & .918 & 1.08 & .001 & .002 & .002 & .006 & .576 & .766 & .818 \\
\midrule
\multirow{6}{*}{SpanDenoise} & Std & .067 & .091 & .105 & .134 & .181 & .220 & .301 & .003 & .006 & .008 & .011 & .025 & .038 & .082 \\
 & Unif & .090 & .118 & .132 & .160 & .204 & .247 & .320 & .005 & .009 & .011 & .016 & .032 & .048 & .087 \\
 & LoRA & .038 & .059 & .065 & .090 & .130 & .160 & .245 & .001 & .002 & .002 & .006 & .014 & .021 & .055 \\
 & FrzLN & .090 & .118 & .131 & .160 & .204 & .246 & .321 & .005 & .009 & .011 & .015 & .030 & .046 & .087 \\
 & FrzInt & .058 & .067 & .069 & .087 & .143 & .182 & .272 & .002 & .002 & .003 & .005 & .018 & .030 & .073 \\
 & EqStep & .306 & .334 & .340 & .370 & .424 & .472 & .540 & .051 & .068 & .068 & .085 & .133 & .168 & .221 \\
\midrule
\multirow{6}{*}{SimCSE} & Std & .099 & .183 & .210 & .289 & .379 & .471 & .657 & .009 & .034 & .046 & .075 & .141 & .194 & .367 \\
 & Unif & .130 & .217 & .247 & .341 & .424 & .501 & .670 & .015 & .044 & .060 & .107 & .181 & .231 & .399 \\
 & LoRA & .039 & .113 & .152 & .224 & .297 & .396 & .570 & .002 & .010 & .022 & .040 & .081 & .132 & .288 \\
 & FrzLN & .131 & .216 & .247 & .334 & .419 & .506 & .672 & .014 & .046 & .063 & .101 & .176 & .233 & .384 \\
 & FrzInt & .076 & .108 & .109 & .137 & .265 & .383 & .626 & .005 & .010 & .009 & .016 & .076 & .127 & .353 \\
 & EqStep & .509 & 1.02 & 1.04 & 1.22 & 1.23 & 1.26 & 1.24 & .436 & .702 & .605 & .775 & .795 & .852 & .822 \\
\midrule
\multirow{6}{*}{BarlowTwins} & Std & .092 & .172 & .224 & .376 & .489 & .537 & .691 & .006 & .024 & .039 & .111 & .188 & .229 & .329 \\
 & Unif & .119 & .209 & .260 & .410 & .518 & .565 & .772 & .008 & .029 & .047 & .122 & .203 & .248 & .402 \\
 & LoRA & .076 & .136 & .182 & .303 & .364 & .407 & .453 & .004 & .013 & .025 & .082 & .108 & .146 & .170 \\
 & FrzLN & .125 & .221 & .272 & .423 & .529 & .575 & .615 & .010 & .033 & .050 & .130 & .206 & .255 & .268 \\
 & FrzInt & .082 & .106 & .118 & .188 & .401 & .467 & .575 & .005 & .007 & .011 & .032 & .141 & .180 & .220 \\
 & EqStep & .376 & .530 & .585 & 1.06 & .861 & .903 & .979 & .156 & .423 & .443 & .651 & 1.00 & 1.00 & 1.00 \\
\bottomrule
\end{tabular}
\end{table}

\begin{table}[ht]
\centering
\caption{Full depth profiles: RoBERTa-base.}
\label{tab:fullprof:roberta-base}
\tiny
\begin{tabular}{llcccccccccccccc}
\toprule
 & & \multicolumn{7}{c}{$\Delta_P$} & \multicolumn{7}{c}{$\Delta_{\text{CKA}}$} \\
\cmidrule(lr){3-9} \cmidrule(lr){10-16}
Obj. & Cond. & 0.10 & 0.25 & 0.40 & 0.60 & 0.75 & 0.90 & 1.00 & 0.10 & 0.25 & 0.40 & 0.60 & 0.75 & 0.90 & 1.00 \\
\midrule
\multirow{5}{*}{MLM} & Std & .082 & .109 & .122 & .186 & .211 & .206 & .245 & .006 & .006 & .007 & .019 & .025 & .021 & .043 \\
 & Unif & .099 & .128 & .145 & .214 & .232 & .227 & .270 & .008 & .008 & .011 & .029 & .031 & .027 & .052 \\
 & LoRA & .073 & .089 & .099 & .182 & .197 & .192 & .212 & .006 & .004 & .006 & .022 & .022 & .022 & .036 \\
 & FrzLN & .100 & .124 & .142 & .202 & .223 & .217 & .260 & .008 & .007 & .010 & .021 & .025 & .022 & .044 \\
 & FrzInt & .058 & .068 & .070 & .116 & .157 & .158 & .198 & .004 & .003 & .002 & .007 & .014 & .013 & .034 \\
\midrule
\multirow{5}{*}{SpanDenoise} & Std & .080 & .110 & .136 & .213 & .236 & .231 & .282 & .006 & .007 & .012 & .034 & .039 & .034 & .065 \\
 & Unif & .099 & .130 & .152 & .229 & .258 & .250 & .295 & .008 & .008 & .012 & .038 & .056 & .048 & .076 \\
 & LoRA & .072 & .090 & .103 & .179 & .191 & .186 & .210 & .005 & .004 & .007 & .024 & .024 & .021 & .039 \\
 & FrzLN & .099 & .129 & .145 & .207 & .239 & .234 & .289 & .008 & .008 & .009 & .020 & .035 & .030 & .064 \\
 & FrzInt & .053 & .064 & .068 & .132 & .172 & .169 & .203 & .003 & .003 & .002 & .009 & .014 & .014 & .028 \\
\midrule
\multirow{5}{*}{SimCSE} & Std & .087 & .178 & .235 & .383 & .457 & .516 & .681 & .005 & .021 & .055 & .181 & .256 & .311 & .459 \\
 & Unif & .105 & .205 & .254 & .396 & .472 & .527 & .676 & .007 & .025 & .048 & .169 & .260 & .320 & .470 \\
 & LoRA & .064 & .175 & .207 & .318 & .393 & .480 & .684 & .002 & .021 & .030 & .063 & .140 & .238 & .470 \\
 & FrzLN & .110 & .211 & .262 & .398 & .467 & .533 & .689 & .007 & .027 & .056 & .172 & .258 & .310 & .471 \\
 & FrzInt & .036 & .044 & .049 & .157 & .312 & .432 & .663 & .001 & .001 & .001 & .018 & .095 & .204 & .440 \\
\midrule
\multirow{5}{*}{BarlowTwins} & Std & .113 & .177 & .214 & .384 & .534 & .573 & .632 & .011 & .021 & .025 & .093 & .206 & .227 & .279 \\
 & Unif & .134 & .210 & .249 & .420 & .581 & .618 & .668 & .013 & .026 & .034 & .121 & .240 & .261 & .310 \\
 & LoRA & .113 & .149 & .182 & .318 & .424 & .457 & .547 & .011 & .012 & .020 & .064 & .162 & .207 & .299 \\
 & FrzLN & .133 & .206 & .245 & .420 & .590 & .618 & .669 & .013 & .024 & .036 & .128 & .274 & .282 & .326 \\
 & FrzInt & .069 & .088 & .091 & .195 & .430 & .503 & .588 & .003 & .004 & .004 & .020 & .132 & .186 & .249 \\
\bottomrule
\end{tabular}
\end{table}

\begin{table}[ht]
\centering
\caption{Full depth profiles: BERT-large.}
\label{tab:fullprof:bert-large}
\tiny
\begin{tabular}{llcccccccccccccc}
\toprule
 & & \multicolumn{7}{c}{$\Delta_P$} & \multicolumn{7}{c}{$\Delta_{\text{CKA}}$} \\
\cmidrule(lr){3-9} \cmidrule(lr){10-16}
Obj. & Cond. & 0.10 & 0.25 & 0.40 & 0.60 & 0.75 & 0.90 & 1.00 & 0.10 & 0.25 & 0.40 & 0.60 & 0.75 & 0.90 & 1.00 \\
\midrule
\multirow{5}{*}{MLM} & Std & .066 & .110 & .123 & .138 & .188 & .359 & .574 & .003 & .009 & .010 & .015 & .027 & .158 & .443 \\
 & Unif & .116 & .167 & .172 & .183 & .234 & .368 & .583 & .010 & .021 & .019 & .028 & .049 & .157 & .447 \\
 & LoRA & .047 & .079 & .088 & .099 & .155 & .340 & .560 & .002 & .004 & .006 & .008 & .017 & .136 & .419 \\
 & FrzLN & .110 & .159 & .172 & .182 & .232 & .384 & .574 & .009 & .017 & .020 & .030 & .051 & .181 & .431 \\
 & FrzInt & .053 & .065 & .061 & .081 & .156 & .335 & .564 & .002 & .002 & .002 & .005 & .018 & .131 & .425 \\
\midrule
\multirow{5}{*}{NSP} & Std & .022 & .104 & .156 & .229 & .323 & .559 & .735 & .001 & .009 & .026 & .044 & .065 & .224 & .497 \\
 & Unif & .047 & .098 & .186 & .403 & .537 & .632 & .733 & .003 & .006 & .030 & .135 & .258 & .351 & 1.00 \\
 & LoRA & .034 & .107 & .158 & .153 & .235 & .490 & .698 & .001 & .012 & .021 & .015 & .041 & .227 & .504 \\
 & FrzLN & .098 & .307 & .451 & .745 & .861 & .911 & .901 & .010 & .081 & .176 & .578 & .693 & .732 & 1.00 \\
 & FrzInt & .012 & .014 & .013 & .064 & .222 & .493 & .671 & .000 & .000 & .000 & .003 & .033 & .233 & .502 \\
\midrule
\multirow{5}{*}{SpanDenoise} & Std & .063 & .100 & .112 & .125 & .197 & .360 & .564 & .003 & .008 & .008 & .011 & .042 & .164 & .426 \\
 & Unif & .115 & .153 & .171 & .174 & .228 & .392 & .573 & .009 & .013 & .021 & .022 & .048 & .191 & .414 \\
 & LoRA & .047 & .076 & .086 & .084 & .132 & .309 & .560 & .002 & .003 & .005 & .004 & .014 & .120 & .438 \\
 & FrzLN & .114 & .157 & .168 & .177 & .236 & .381 & .611 & .009 & .020 & .017 & .027 & .060 & .171 & .483 \\
 & FrzInt & .051 & .061 & .057 & .078 & .157 & .326 & .579 & .002 & .002 & .002 & .005 & .024 & .125 & .467 \\
\midrule
\multirow{5}{*}{SimCSE} & Std & .070 & .150 & .201 & .279 & .355 & .558 & .792 & .005 & .017 & .040 & .097 & .141 & .338 & .657 \\
 & Unif & .162 & .248 & .306 & .385 & .468 & .634 & .799 & .029 & .058 & .102 & .181 & .260 & .441 & .677 \\
 & LoRA & .030 & .126 & .173 & .214 & .289 & .489 & .766 & .001 & .014 & .035 & .053 & .098 & .246 & .633 \\
 & FrzLN & .130 & .215 & .271 & .356 & .430 & .581 & .798 & .013 & .034 & .077 & .161 & .202 & .341 & .675 \\
 & FrzInt & .035 & .039 & .037 & .145 & .323 & .515 & .779 & .001 & .001 & .001 & .022 & .133 & .297 & .653 \\
\midrule
\multirow{5}{*}{BarlowTwins} & Std & .083 & .151 & .192 & .283 & .422 & .563 & .720 & .005 & .017 & .026 & .095 & .199 & .297 & .559 \\
 & Unif & .142 & .220 & .269 & .350 & .472 & .627 & .757 & .019 & .034 & .078 & .143 & .230 & .354 & .613 \\
 & LoRA & .074 & .136 & .182 & .264 & .346 & .489 & .643 & .005 & .013 & .028 & .083 & .117 & .250 & .509 \\
 & FrzLN & .137 & .219 & .259 & .353 & .485 & .635 & .760 & .015 & .031 & .061 & .158 & .234 & .371 & .596 \\
 & FrzInt & .066 & .087 & .090 & .165 & .385 & .552 & .704 & .005 & .006 & .006 & .027 & .178 & .296 & .543 \\
\bottomrule
\end{tabular}
\end{table}

\begin{table}[ht]
\centering
\caption{Full depth profiles: Pythia-160m.}
\label{tab:fullprof:pythia-160m}
\tiny
\begin{tabular}{llcccccccccccccc}
\toprule
 & & \multicolumn{7}{c}{$\Delta_P$} & \multicolumn{7}{c}{$\Delta_{\text{CKA}}$} \\
\cmidrule(lr){3-9} \cmidrule(lr){10-16}
Obj. & Cond. & 0.10 & 0.25 & 0.40 & 0.60 & 0.75 & 0.90 & 1.00 & 0.10 & 0.25 & 0.40 & 0.60 & 0.75 & 0.90 & 1.00 \\
\midrule
\multirow{6}{*}{CausalLM} & Std & .114 & .120 & .088 & .094 & .124 & .160 & .278 & .012 & .007 & .006 & .006 & .008 & .010 & .070 \\
 & Unif & .123 & .173 & .109 & .121 & .157 & .195 & .285 & .013 & .015 & .009 & .009 & .012 & .013 & .068 \\
 & LoRA & .035 & .035 & .027 & .032 & .051 & .076 & .203 & .002 & .000 & .001 & .001 & .001 & .001 & .034 \\
 & FrzLN & .123 & .643 & .298 & .339 & .427 & .511 & .294 & .013 & .226 & .024 & .026 & .040 & .058 & .078 \\
 & FrzInt & .092 & .079 & .071 & .072 & .088 & .117 & .239 & .007 & .004 & .005 & .005 & .005 & .006 & .053 \\
 & EqStep & .373 & .805 & .956 & .881 & .767 & .702 & .479 & .098 & .484 & .441 & .337 & .315 & .296 & .144 \\
\midrule
\multirow{6}{*}{CausalSpan} & Std & .094 & .155 & .131 & .149 & .174 & .210 & .313 & .005 & .015 & .014 & .014 & .015 & .017 & .091 \\
 & Unif & .137 & .178 & .132 & .167 & .209 & .265 & .332 & .021 & .012 & .008 & .008 & .009 & .013 & .088 \\
 & LoRA & .051 & .042 & .027 & .031 & .047 & .065 & .200 & .002 & .001 & .001 & .001 & .001 & .001 & .034 \\
 & FrzLN & .135 & .561 & .395 & .447 & .543 & .587 & .331 & .017 & .164 & .088 & .089 & .105 & .112 & .086 \\
 & FrzInt & .085 & .131 & .083 & .089 & .124 & .166 & .280 & .005 & .015 & .006 & .007 & .007 & .009 & .079 \\
 & EqStep & .705 & .969 & 1.10 & .949 & .953 & .935 & .792 & .251 & .748 & .739 & .511 & .560 & .575 & .364 \\
\midrule
\multirow{6}{*}{SimCSE} & Std & .132 & .979 & 1.18 & 1.17 & 1.17 & 1.14 & .571 & .019 & .859 & .919 & .911 & .916 & .912 & .351 \\
 & Unif & .245 & .903 & 1.19 & 1.13 & 1.12 & 1.08 & .632 & .075 & .751 & .937 & .870 & .854 & .838 & .442 \\
 & LoRA & .009 & .018 & .015 & .047 & .068 & .115 & .478 & .000 & .000 & .000 & .001 & .001 & .003 & .301 \\
 & FrzLN & .299 & 1.11 & 1.24 & 1.21 & 1.21 & 1.19 & .725 & .116 & .932 & .956 & .938 & .946 & .945 & .552 \\
 & FrzInt & .046 & .032 & .028 & .036 & .132 & .166 & .394 & .002 & .001 & .001 & .001 & .004 & .003 & .161 \\
 & EqStep & 1.16 & 1.20 & 1.12 & 1.14 & 1.12 & 1.08 & 1.08 & .800 & .914 & .861 & .840 & .802 & .765 & .717 \\
\midrule
\multirow{6}{*}{BarlowTwins} & Std & .205 & .913 & .952 & .956 & .973 & .958 & .537 & .051 & .716 & .708 & .706 & .735 & .728 & .307 \\
 & Unif & .310 & .682 & .915 & .926 & .945 & .969 & .539 & .112 & .415 & .658 & .636 & .672 & .721 & .304 \\
 & LoRA & .024 & .074 & .056 & .069 & .090 & .126 & .417 & .001 & .005 & .003 & .004 & .004 & .006 & .171 \\
 & FrzLN & .274 & 1.00 & 1.12 & 1.12 & 1.12 & 1.11 & .514 & .065 & .812 & .872 & .873 & .878 & .879 & .271 \\
 & FrzInt & .087 & .093 & .131 & .148 & .176 & .215 & .414 & .005 & .007 & .017 & .015 & .018 & .028 & .158 \\
 & EqStep & 1.17 & 1.14 & 1.18 & 1.08 & 1.06 & 1.05 & .915 & .834 & .868 & .877 & .728 & .698 & .702 & .664 \\
\bottomrule
\end{tabular}
\end{table}

\begin{table}[ht]
\centering
\caption{Full depth profiles: OPT-125m.}
\label{tab:fullprof:opt-125m}
\tiny
\begin{tabular}{llcccccccccccccc}
\toprule
 & & \multicolumn{7}{c}{$\Delta_P$} & \multicolumn{7}{c}{$\Delta_{\text{CKA}}$} \\
\cmidrule(lr){3-9} \cmidrule(lr){10-16}
Obj. & Cond. & 0.10 & 0.25 & 0.40 & 0.60 & 0.75 & 0.90 & 1.00 & 0.10 & 0.25 & 0.40 & 0.60 & 0.75 & 0.90 & 1.00 \\
\midrule
\multirow{6}{*}{CausalLM} & Std & .007 & .022 & .023 & .038 & .057 & .075 & .278 & .000 & .000 & .000 & .000 & .000 & .000 & .081 \\
 & Unif & .007 & .036 & .036 & .057 & .081 & .104 & .285 & .000 & .000 & .000 & .000 & .000 & .000 & .083 \\
 & LoRA & .010 & .006 & .006 & .011 & .020 & .029 & .236 & .000 & .000 & .000 & .000 & .000 & .000 & .050 \\
 & FrzLN & .008 & .039 & .039 & .060 & .085 & .110 & .287 & .000 & .000 & .000 & .000 & .000 & .000 & .088 \\
 & FrzInt & .013 & .006 & .006 & .011 & .019 & .030 & .255 & .000 & .000 & .000 & .000 & .000 & .000 & .068 \\
 & EqStep & .092 & .089 & .093 & .156 & .214 & .257 & .378 & .000 & .000 & .000 & .000 & .000 & .001 & .114 \\
\midrule
\multirow{6}{*}{CausalSpan} & Std & .012 & .006 & .007 & .015 & .024 & .035 & .278 & .000 & .000 & .000 & .000 & .000 & .000 & .074 \\
 & Unif & .011 & .007 & .008 & .018 & .029 & .040 & .295 & .000 & .000 & .000 & .000 & .000 & .000 & .080 \\
 & LoRA & .011 & .005 & .006 & .010 & .019 & .027 & .228 & .000 & .000 & .000 & .000 & .000 & .000 & .048 \\
 & FrzLN & .011 & .007 & .008 & .018 & .028 & .040 & .295 & .000 & .000 & .000 & .000 & .000 & .000 & .080 \\
 & FrzInt & .011 & .005 & .005 & .010 & .019 & .029 & .251 & .000 & .000 & .000 & .000 & .000 & .000 & .060 \\
 & EqStep & .141 & .141 & .153 & .260 & .336 & .395 & .473 & .000 & .000 & .000 & .001 & .002 & .004 & .130 \\
\midrule
\multirow{6}{*}{SimCSE} & Std & .009 & .008 & .009 & .021 & .033 & .060 & .485 & .000 & .000 & .000 & .000 & .000 & .000 & .204 \\
 & Unif & .009 & .010 & .011 & .024 & .037 & .057 & .456 & .000 & .000 & .000 & .000 & .000 & .000 & .157 \\
 & LoRA & .003 & .003 & .003 & .009 & .016 & .030 & .370 & .000 & .000 & .000 & .000 & .000 & .000 & .103 \\
 & FrzLN & .010 & .009 & .010 & .026 & .042 & .082 & .434 & .000 & .000 & .000 & .000 & .000 & .000 & .150 \\
 & FrzInt & .003 & .002 & .002 & .006 & .015 & .040 & .406 & .000 & .000 & .000 & .000 & .000 & .000 & .123 \\
 & EqStep & 1.19 & 1.13 & 1.12 & 1.16 & 1.16 & 1.16 & .786 & .818 & .645 & .642 & .682 & .710 & .723 & .583 \\
\midrule
\multirow{6}{*}{BarlowTwins} & Std & .013 & .009 & .011 & .032 & .051 & .066 & .433 & .000 & .000 & .000 & .000 & .000 & .000 & .148 \\
 & Unif & .012 & .011 & .014 & .043 & .070 & .090 & .501 & .000 & .000 & .000 & .000 & .000 & .000 & .197 \\
 & LoRA & .009 & .007 & .008 & .025 & .038 & .053 & .293 & .000 & .000 & .000 & .000 & .000 & .000 & .074 \\
 & FrzLN & .014 & .011 & .014 & .039 & .065 & .085 & .491 & .000 & .000 & .000 & .000 & .000 & .000 & .194 \\
 & FrzInt & .016 & .007 & .007 & .012 & .029 & .054 & .363 & .000 & .000 & .000 & .000 & .000 & .000 & .101 \\
 & EqStep & .712 & .506 & .513 & .627 & .797 & .911 & .847 & .134 & .102 & .101 & .099 & .130 & .216 & .582 \\
\bottomrule
\end{tabular}
\end{table}

\begin{table}[ht]
\centering
\caption{Full depth profiles: GPT-2 medium.}
\label{tab:fullprof:gpt2-medium}
\tiny
\begin{tabular}{llcccccccccccccc}
\toprule
 & & \multicolumn{7}{c}{$\Delta_P$} & \multicolumn{7}{c}{$\Delta_{\text{CKA}}$} \\
\cmidrule(lr){3-9} \cmidrule(lr){10-16}
Obj. & Cond. & 0.10 & 0.25 & 0.40 & 0.60 & 0.75 & 0.90 & 1.00 & 0.10 & 0.25 & 0.40 & 0.60 & 0.75 & 0.90 & 1.00 \\
\midrule
\multirow{6}{*}{CausalLM} & Std & .081 & .045 & .049 & .068 & .105 & .147 & .795 & .004 & .001 & .001 & .001 & .001 & .001 & .596 \\
 & Unif & .101 & .063 & .071 & .095 & .137 & .182 & .813 & .008 & .001 & .001 & .001 & .001 & .002 & .614 \\
 & LoRA & .083 & .027 & .028 & .040 & .071 & .104 & .574 & .005 & .000 & .000 & .000 & .000 & .001 & .323 \\
 & FrzLN & .102 & .058 & .064 & .086 & .125 & .167 & .817 & .008 & .001 & .001 & .001 & .001 & .002 & .618 \\
 & FrzInt & .091 & .016 & .016 & .023 & .056 & .089 & .756 & .006 & .000 & .000 & .000 & .000 & .001 & .555 \\
 & EqStep & .254 & .159 & .179 & .212 & .261 & .308 & 1.07 & .044 & .015 & .018 & .020 & .020 & .025 & .793 \\
\midrule
\multirow{6}{*}{CausalSpan} & Std & .081 & .040 & .044 & .059 & .094 & .126 & .691 & .005 & .000 & .001 & .001 & .001 & .001 & .455 \\
 & Unif & .101 & .056 & .064 & .086 & .124 & .158 & .702 & .009 & .001 & .001 & .001 & .001 & .001 & .462 \\
 & LoRA & .080 & .021 & .022 & .030 & .059 & .088 & .586 & .006 & .000 & .000 & .000 & .000 & .001 & .340 \\
 & FrzLN & .100 & .055 & .062 & .082 & .118 & .151 & .701 & .008 & .001 & .001 & .001 & .001 & .001 & .460 \\
 & FrzInt & .079 & .014 & .014 & .023 & .065 & .096 & .655 & .004 & .000 & .000 & .000 & .000 & .000 & .413 \\
 & EqStep & .328 & .217 & .261 & .322 & .370 & .402 & 1.09 & .061 & .015 & .020 & .025 & .030 & .036 & .829 \\
\midrule
\multirow{6}{*}{SimCSE} & Std & .100 & .033 & .044 & .081 & .230 & .361 & 1.12 & .004 & .000 & .000 & .000 & .001 & .006 & .874 \\
 & Unif & .138 & .046 & .051 & .094 & .227 & .337 & 1.13 & .009 & .000 & .000 & .001 & .002 & .005 & .888 \\
 & LoRA & .109 & .042 & .052 & .101 & .296 & .448 & 1.12 & .010 & .000 & .000 & .000 & .003 & .018 & .906 \\
 & FrzLN & .128 & .066 & .078 & .139 & .316 & .438 & 1.14 & .007 & .001 & .001 & .001 & .004 & .012 & .906 \\
 & FrzInt & .036 & .005 & .005 & .018 & .195 & .357 & 1.11 & .001 & .000 & .000 & .000 & .001 & .007 & .866 \\
 & EqStep & 1.07 & 1.02 & 1.03 & 1.01 & .973 & .967 & 1.13 & .653 & .677 & .710 & .679 & .633 & .620 & .836 \\
\midrule
\multirow{6}{*}{BarlowTwins} & Std & .108 & .030 & .042 & .072 & .148 & .227 & .944 & .006 & .000 & .000 & .000 & .001 & .003 & .723 \\
 & Unif & .137 & .041 & .055 & .088 & .165 & .237 & 1.00 & .010 & .000 & .000 & .001 & .001 & .003 & .781 \\
 & LoRA & .136 & .027 & .039 & .067 & .136 & .201 & 1.02 & .014 & .000 & .000 & .000 & .001 & .003 & .807 \\
 & FrzLN & .150 & .043 & .063 & .099 & .177 & .241 & 1.00 & .011 & .000 & .001 & .001 & .002 & .004 & .776 \\
 & FrzInt & .108 & .020 & .019 & .025 & .122 & .234 & .956 & .007 & .000 & .000 & .000 & .001 & .003 & .729 \\
 & EqStep & .867 & .786 & .814 & .893 & 1.01 & 1.09 & 1.16 & .617 & .432 & .423 & .437 & .573 & .772 & .889 \\
\bottomrule
\end{tabular}
\end{table}

\begin{table}[ht]
\centering
\caption{Full depth profiles: Mamba-130m.}
\label{tab:fullprof:mamba-130m}
\tiny
\begin{tabular}{llcccccccccccccc}
\toprule
 & & \multicolumn{7}{c}{$\Delta_P$} & \multicolumn{7}{c}{$\Delta_{\text{CKA}}$} \\
\cmidrule(lr){3-9} \cmidrule(lr){10-16}
Obj. & Cond. & 0.10 & 0.25 & 0.40 & 0.60 & 0.75 & 0.90 & 1.00 & 0.10 & 0.25 & 0.40 & 0.60 & 0.75 & 0.90 & 1.00 \\
\midrule
\multirow{5}{*}{CausalLM} & Std & .024 & .088 & .097 & .133 & .156 & .203 & .507 & .000 & .005 & .007 & .014 & .021 & .036 & .296 \\
 & Unif & .032 & .101 & .110 & .149 & .171 & .216 & .564 & .001 & .007 & .008 & .018 & .025 & .038 & .377 \\
 & LoRA & .021 & .091 & .099 & .137 & .163 & .209 & .509 & .000 & .006 & .007 & .015 & .022 & .036 & .277 \\
 & FrzLN & .032 & .101 & .110 & .148 & .170 & .215 & .565 & .001 & .006 & .008 & .018 & .025 & .038 & .379 \\
 & FrzInt & .011 & .035 & .032 & .063 & .097 & .157 & .485 & .000 & .001 & .001 & .003 & .008 & .024 & .297 \\
\midrule
\multirow{5}{*}{CausalSpan} & Std & .027 & .088 & .091 & .127 & .149 & .198 & .546 & .001 & .006 & .006 & .012 & .017 & .034 & .348 \\
 & Unif & .037 & .100 & .108 & .148 & .171 & .220 & .596 & .001 & .007 & .007 & .016 & .024 & .041 & .413 \\
 & LoRA & .026 & .079 & .080 & .112 & .137 & .182 & .481 & .001 & .005 & .004 & .010 & .016 & .030 & .259 \\
 & FrzLN & .037 & .100 & .107 & .148 & .171 & .219 & .601 & .001 & .007 & .007 & .016 & .023 & .040 & .425 \\
 & FrzInt & .009 & .020 & .018 & .049 & .094 & .149 & .523 & .000 & .000 & .000 & .002 & .008 & .025 & .337 \\
\midrule
\multirow{5}{*}{SimCSE} & Std & .026 & .038 & .042 & .065 & .127 & .416 & .646 & .001 & .001 & .001 & .003 & .019 & .158 & .348 \\
 & Unif & .043 & .115 & .168 & .278 & .361 & .474 & .818 & .002 & .009 & .026 & .133 & .236 & .313 & .599 \\
 & LoRA & .006 & .021 & .027 & .059 & .127 & .210 & .445 & .000 & .000 & .000 & .003 & .022 & .058 & .204 \\
 & FrzLN & .037 & .069 & .085 & .220 & .322 & .457 & .806 & .001 & .003 & .005 & .074 & .179 & .298 & .582 \\
 & FrzInt & .001 & .002 & .002 & .019 & .077 & .434 & .658 & .000 & .000 & .000 & .000 & .005 & .181 & .347 \\
\midrule
\multirow{5}{*}{BarlowTwins} & Std & .028 & .076 & .111 & .213 & .302 & .442 & .771 & .001 & .003 & .008 & .057 & .137 & .224 & .453 \\
 & Unif & .073 & .316 & .329 & .399 & .456 & .559 & .886 & .004 & .170 & .159 & .206 & .299 & .324 & .599 \\
 & LoRA & .015 & .062 & .089 & .187 & .274 & .386 & .795 & .000 & .002 & .006 & .047 & .125 & .203 & .669 \\
 & FrzLN & .237 & .280 & .331 & .433 & .483 & .577 & .894 & .125 & .056 & .099 & .189 & .257 & .305 & .557 \\
 & FrzInt & .004 & .007 & .006 & .042 & .163 & .332 & .707 & .000 & .000 & .000 & .001 & .025 & .102 & .369 \\
\bottomrule
\end{tabular}
\end{table}

\begin{table}[ht]
\centering
\caption{Full depth profiles: RWKV-4-169m.}
\label{tab:fullprof:rwkv-169m}
\tiny
\begin{tabular}{llcccccccccccccc}
\toprule
 & & \multicolumn{7}{c}{$\Delta_P$} & \multicolumn{7}{c}{$\Delta_{\text{CKA}}$} \\
\cmidrule(lr){3-9} \cmidrule(lr){10-16}
Obj. & Cond. & 0.10 & 0.25 & 0.40 & 0.60 & 0.75 & 0.90 & 1.00 & 0.10 & 0.25 & 0.40 & 0.60 & 0.75 & 0.90 & 1.00 \\
\midrule
\multirow{5}{*}{CausalLM} & Std & .036 & .072 & .093 & .177 & .250 & .287 & .454 & .002 & .007 & .010 & .045 & .087 & .110 & .285 \\
 & Unif & .041 & .081 & .101 & .186 & .261 & .296 & .447 & .003 & .009 & .012 & .050 & .096 & .118 & .270 \\
 & LoRA & .036 & .075 & .097 & .170 & .243 & .305 & .811 & .002 & .008 & .011 & .040 & .074 & .117 & .712 \\
 & FrzLN & .035 & .077 & .097 & .181 & .253 & .287 & .440 & .002 & .008 & .011 & .047 & .087 & .105 & .258 \\
 & FrzInt & .010 & .029 & .031 & .074 & .177 & .217 & .494 & .000 & .002 & .001 & .007 & .054 & .076 & .376 \\
\midrule
\multirow{5}{*}{CausalSpan} & Std & .033 & .067 & .085 & .155 & .210 & .239 & .654 & .002 & .005 & .007 & .034 & .058 & .070 & .575 \\
 & Unif & .037 & .080 & .097 & .171 & .221 & .248 & .648 & .003 & .007 & .010 & .044 & .066 & .076 & .567 \\
 & LoRA & .018 & .058 & .083 & .151 & .234 & .269 & .654 & .001 & .004 & .009 & .031 & .086 & .107 & .600 \\
 & FrzLN & .029 & .075 & .094 & .167 & .220 & .248 & .649 & .002 & .006 & .010 & .043 & .067 & .077 & .565 \\
 & FrzInt & .006 & .015 & .018 & .056 & .158 & .198 & .722 & .000 & .000 & .000 & .003 & .041 & .061 & .624 \\
\midrule
\multirow{5}{*}{SimCSE} & Std & .019 & .050 & .142 & .433 & .550 & .623 & .940 & .001 & .003 & .036 & .323 & .474 & .499 & .867 \\
 & Unif & .026 & .063 & .152 & .488 & .584 & .657 & .944 & .001 & .005 & .037 & .395 & .491 & .513 & .846 \\
 & LoRA & .010 & .048 & .132 & .432 & .579 & .613 & .902 & .000 & .002 & .029 & .355 & .512 & .476 & .828 \\
 & FrzLN & .016 & .069 & .180 & .440 & .561 & .633 & .938 & .000 & .004 & .056 & .347 & .487 & .513 & .881 \\
 & FrzInt & .001 & .002 & .002 & .020 & .104 & .180 & .635 & .000 & .000 & .000 & .001 & .015 & .030 & .468 \\
\midrule
\multirow{5}{*}{BarlowTwins} & Std & .058 & .104 & .133 & .242 & .408 & .482 & .726 & .005 & .015 & .032 & .087 & .233 & .287 & .575 \\
 & Unif & .048 & .127 & .153 & .966 & .836 & .742 & .949 & .003 & .022 & .030 & .911 & .862 & .725 & .875 \\
 & LoRA & .012 & .037 & .061 & .198 & .354 & .516 & .804 & .000 & .002 & .006 & .079 & .191 & .320 & .626 \\
 & FrzLN & .046 & .102 & .151 & .279 & .457 & .497 & .676 & .003 & .013 & .042 & .124 & .336 & .354 & .511 \\
 & FrzInt & .004 & .006 & .007 & .073 & .274 & .392 & .674 & .000 & .000 & .000 & .010 & .118 & .196 & .467 \\
\bottomrule
\end{tabular}
\end{table}

\begin{table}[ht]
\centering
\caption{Full depth profiles: CodeGen-350M.}
\label{tab:fullprof:codegen-350m}
\tiny
\begin{tabular}{llcccccccccccccc}
\toprule
 & & \multicolumn{7}{c}{$\Delta_P$} & \multicolumn{7}{c}{$\Delta_{\text{CKA}}$} \\
\cmidrule(lr){3-9} \cmidrule(lr){10-16}
Obj. & Cond. & 0.10 & 0.25 & 0.40 & 0.60 & 0.75 & 0.90 & 1.00 & 0.10 & 0.25 & 0.40 & 0.60 & 0.75 & 0.90 & 1.00 \\
\midrule
\multirow{6}{*}{CausalLM} & Std & .155 & .164 & .174 & .201 & .254 & .320 & .430 & .015 & .018 & .018 & .018 & .021 & .030 & .194 \\
 & Unif & .179 & .210 & .221 & .252 & .303 & .361 & .446 & .017 & .026 & .026 & .027 & .033 & .045 & .204 \\
 & LoRA & .074 & .090 & .098 & .113 & .144 & .192 & .358 & .006 & .007 & .007 & .006 & .006 & .007 & .134 \\
 & FrzLN & .179 & .202 & .213 & .244 & .295 & .355 & .446 & .016 & .024 & .024 & .024 & .030 & .042 & .203 \\
 & FrzInt & .142 & .142 & .143 & .150 & .194 & .260 & .398 & .013 & .018 & .018 & .018 & .019 & .023 & .171 \\
 & EqStep & .368 & .413 & .446 & .497 & .525 & .541 & .480 & .093 & .073 & .071 & .080 & .100 & .135 & .223 \\
\midrule
\multirow{6}{*}{CausalSpan} & Std & .135 & .202 & .229 & .267 & .325 & .381 & .443 & .011 & .021 & .021 & .022 & .029 & .048 & .189 \\
 & Unif & .193 & .267 & .303 & .349 & .396 & .445 & .464 & .022 & .020 & .021 & .026 & .038 & .071 & .202 \\
 & LoRA & .033 & .161 & .160 & .172 & .206 & .261 & .348 & .001 & .023 & .022 & .021 & .021 & .025 & .134 \\
 & FrzLN & .197 & .282 & .314 & .356 & .400 & .446 & .463 & .023 & .034 & .034 & .037 & .048 & .079 & .201 \\
 & FrzInt & .106 & .110 & .110 & .124 & .184 & .251 & .380 & .007 & .010 & .010 & .009 & .010 & .015 & .157 \\
 & EqStep & .431 & .577 & .611 & .647 & .666 & .669 & .574 & .082 & .118 & .127 & .156 & .205 & .280 & .247 \\
\midrule
\multirow{6}{*}{SimCSE} & Std & .054 & .167 & .189 & .181 & .223 & .292 & .561 & .003 & .016 & .017 & .018 & .020 & .023 & .277 \\
 & Unif & .493 & .888 & .891 & .899 & .900 & .911 & .676 & .268 & .650 & .653 & .654 & .647 & .644 & .466 \\
 & LoRA & .008 & .024 & .053 & .113 & .186 & .277 & .519 & .000 & .000 & .000 & .001 & .002 & .004 & .237 \\
 & FrzLN & .535 & 1.14 & 1.09 & 1.06 & 1.05 & 1.05 & .708 & .197 & .932 & .900 & .887 & .887 & .890 & .540 \\
 & FrzInt & .032 & .017 & .017 & .043 & .148 & .233 & .514 & .001 & .000 & .000 & .000 & .001 & .003 & .230 \\
 & EqStep & 1.08 & .967 & .956 & .962 & .943 & .930 & .898 & .623 & .633 & .613 & .631 & .604 & .589 & .627 \\
\midrule
\multirow{6}{*}{BarlowTwins} & Std & .181 & .564 & .565 & .575 & .579 & .577 & .568 & .035 & .321 & .320 & .320 & .321 & .318 & .362 \\
 & Unif & .400 & 1.04 & 1.05 & 1.06 & 1.05 & 1.04 & .646 & .229 & .820 & .821 & .822 & .819 & .812 & .568 \\
 & LoRA & .015 & .030 & .048 & .127 & .212 & .304 & .494 & .000 & .000 & .000 & .001 & .002 & .004 & .247 \\
 & FrzLN & .878 & .549 & .559 & .577 & .606 & .645 & .765 & .655 & .241 & .242 & .244 & .246 & .249 & .651 \\
 & FrzInt & .120 & .071 & .072 & .095 & .170 & .225 & .479 & .017 & .004 & .004 & .004 & .005 & .008 & .242 \\
 & EqStep & .925 & 1.01 & .997 & .970 & .932 & .940 & .750 & .535 & .774 & .763 & .747 & .716 & .747 & .578 \\
\bottomrule
\end{tabular}
\end{table}

\end{document}